\documentclass[final]{cvpr}

\usepackage{times}
\usepackage{epsfig}
\usepackage{graphicx}
\usepackage{amsmath}
\usepackage{amssymb}
\usepackage[dvipsnames]{xcolor}

\usepackage{booktabs}

\usepackage{makecell}

\usepackage{array, multirow}
\newcolumntype{C}[1]{>{\centering\let\newline\\\arraybackslash\hspace{0pt}}m{#1}}

\usepackage[pagebackref=true,breaklinks=true,colorlinks,bookmarks=false]{hyperref}

\def\cvprPaperID{7447} 
\def\confYear{CVPR 2021}

\newcommand{\norm}{\mathcal{N}}

\newcommand{\parsection}[1]{\vspace{0mm}\noindent\textbf{#1}~}

\begin{document}

\title{DeFlow: Learning Complex Image Degradations from Unpaired Data \\with Conditional Flows}

\author{Valentin Wolf \qquad Andreas Lugmayr \qquad
    Martin Danelljan \qquad Luc Van Gool \qquad Radu Timofte\\
    \resizebox{0.96\linewidth}{!}{%
        \centering
        {\tt\small vawolf@ethz.ch} \qquad { \tt\small \{andreas.lugmayr, martin.danelljan, vangool, radu.timofte\}@vision.ee.ethz.ch }
    }\\
    Computer Vision Lab, ETH Zurich, Switzerland\\
    
}

\maketitle
\thispagestyle{empty}

\global\csname @topnum\endcsname 0
\global\csname @botnum\endcsname 0
\begin{abstract}
The difficulty of obtaining paired data remains a major bottleneck for learning image restoration and enhancement models for real-world applications. Current strategies aim to synthesize realistic training data by modeling noise and degradations that appear in real-world settings.  We propose DeFlow, a method for learning stochastic image degradations from unpaired data. Our approach is based on a novel unpaired learning formulation for conditional normalizing flows. We model the degradation process in the latent space of a shared flow encoder-decoder network. This allows us to learn the conditional distribution of a noisy image given the clean input by solely minimizing the negative log-likelihood of the marginal distributions. We validate our DeFlow formulation on the task of joint image restoration and super-resolution. The models trained with the synthetic data generated by DeFlow outperform previous learnable approaches on three recent datasets. Code and trained models are available at: \url{https://github.com/volflow/DeFlow}
\end{abstract}

\section{Introduction}

\begin{figure}
\includegraphics[width=\columnwidth]{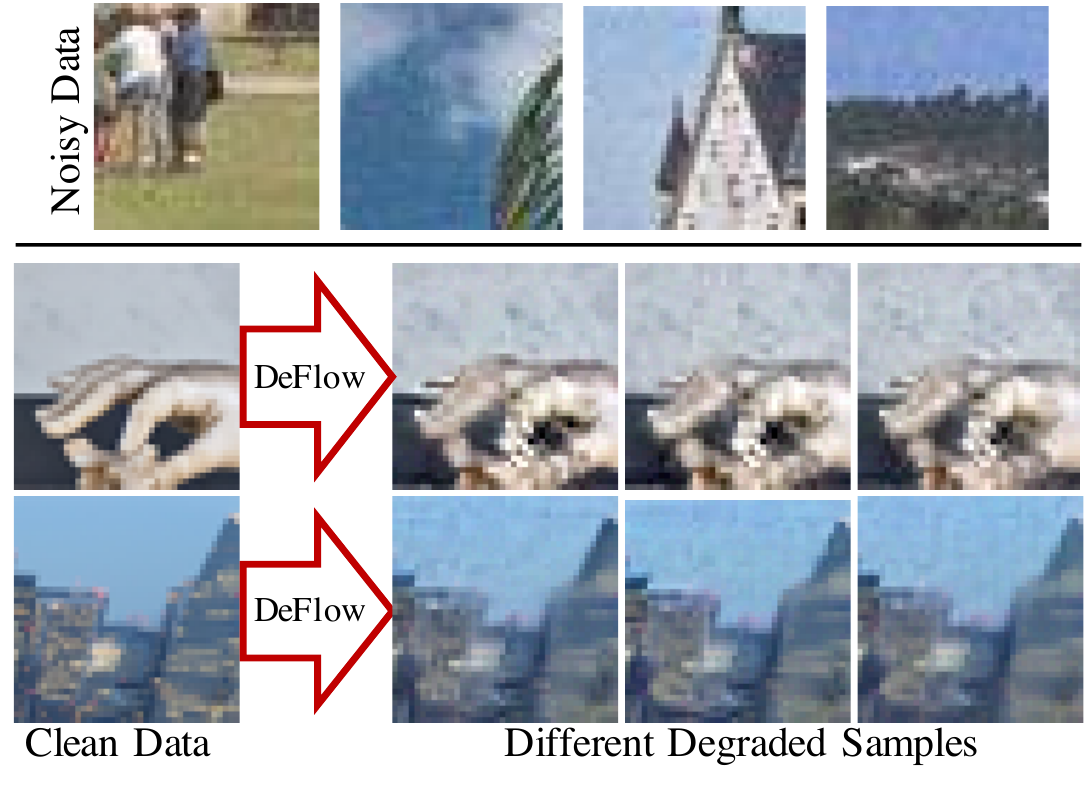}%
\vspace{-2.5mm}
\caption{\textbf{DeFlow} is able to learn complex image degradation processes from unpaired training data. Our approach can sample different degraded versions of a clean input image (bottom) that faithfully resemble the noise of the real data (top).}
\label{fig:opening}\vspace{-5mm}
\end{figure}

Deep learning based methods have demonstrated astonishing performance for image restoration and enhancement when large quantities of paired training data are available.
However, for many real-world applications, obtaining paired data remains a major bottleneck.
For instance, in real-world super-resolution \cite{lugmayrICCVW2019, Cai_2019_CVPR_Workshops, cai2019toward} and denoising \cite{SIDD_2018_CVPR, Abdelhamed_2019_CVPR_Workshops}, collecting paired data is cumbersome and expensive, requiring careful setups and procedures that are difficult to scale. Moreover, such data is often limited to certain scenes and contains substantial misalignment issues. In many settings, including enhancement of existing image collections or restoration of historic photographs, the collection of paired data is even impossible.

To tackle this fundamental problem, one promising direction is to generate paired training data by applying synthesized degradations and noise to high-quality images. The degraded image then has a high-quality ground-truth, allowing effective supervised learning techniques to be applied directly to the synthesized pairs. However, in most practical applications the degradation process is \emph{unknown}. It generally constitutes a complex combination of sensor noise, compression, and post-processing artifacts.
Modeling the degradation process by hand is therefore a highly challenging problem, calling for learnable alternatives.

Since paired data is unavailable, learning the degradation process requires \emph{unpaired} or \emph{unsupervised} techniques. Several approaches resort to hand-crafted strategies tailored to specific types of degradations \cite{Ji_2020_CVPR_Workshops}.
Existing learnable solutions mostly adopt generative adversarial networks (GANs) with cycle-consistency constraints \cite{zhu2017unpaired,lugmayrICCVW2019,bulat2018learn} or domain-aware adversarial objectives \cite{Fritsche19Frequency, wei2020unsupervised, bell2019blindKernelGAN} for unpaired training. However, these approaches require careful tuning of several losses. Moreover, cycle-consistency is a weak constraint that easily leads to changes in color and content \cite{chu2017cyclegan}. Importantly, the aforementioned works rely on fully deterministic mappings, completely ignoring the fundamental stochasticity of natural degradations and noise. In this work, we therefore take a radically different approach.

We propose \emph{DeFlow}: a novel conditional normalizing flow based method for learning degradations from unpaired data. \emph{DeFlow} models the conditional distribution $p(y|x)$ of a degraded image $y$ given its clean counterpart $x$. As shown in Fig.~\ref{fig:opening}, this allows us to sample multiple degraded versions $y$ of any clean image $x$, which closely resemble the characteristics of the unknown degradations.
However, conventional conditional flow models \cite{winkler2019learningConditionalFlows, srflow, Ardizzone19cINNcolor, Abdelhamed2019NoiseFlow} require sample pairs $(x,y)$ for supervised training.
We therefore propose a novel formulation for conditional flows, capable of unpaired learning. Specifically, we treat the unpaired setting as the problem of learning the conditional distribution $p(y|x)$ from observations of the marginals $p(x)$ and $p(y)$. By modeling both domains $x$ and $y$ in the latent space of a joint flow network, we ensure sufficient constraints for effective unpaired learning while preserving flexibility for accurate modeling of $p(y|x)$. We additionally introduce a method for conditioning the flow on domain invariant information derived from either $x$ or $y$ to further facilitate the learning problem. 

We apply our DeFlow formulation to the problem of joint image restoration and super-resolution in the real-world setting. DeFlow is tasked with learning complex image degradations, which are then used to synthesize training data for a baseline super-resolution model. We perform comprehensive experiments and analysis on the AIM2019 \cite{AIM2019RWSRchallenge} and NTIRE2020 \cite{ntire2020RWSR} real-world super-resolution challenge datasets. Our approach sets a new state-of-the-art among learning-based approaches by outperforming GAN-based alternatives for generating image degradations from unpaired data on three datasets.

\section{Related Work}

\parsection{Learning degradations from unpaired data}
Realistic noise modeling and generation is a long-standing problem in Computer Vision research. 
The direction of finding learning-based solutions capable of utilizing unpaired data has received growing interest. One line of research employs generative adversarial networks (GANs) \cite{Goodfellow14GAN}. To learn from unpaired data, either cycle-consistency losses \cite{lugmayrICCVW2019, bulat2018learn} or domain-based adversarial losses \cite{Fritsche19Frequency, wei2020unsupervised, bell2019blindKernelGAN} are employed. Yet, these approaches suffer from convergence and mode collapse issues, requiring elaborate fine-tuning of their losses. Importantly, such methods learn a deterministic mapping, ignoring the stochasticity of degradations.

Other works \cite{krull2018noise2void, prakash2020fully, laine2019highquality, wu2020unpaired} learn unsupervised denoising models based on the assumption of spatially uncorrelated (\ie white) noise. However, this assumption does not apply to more complex degradations, which have substantial spatial correlation due to \eg compression or post-processing artifacts. Our approach exploits fundamentally different constraints to allow for unpaired learning in this more challenging setting.
Recently Abdelhamed \etal \cite{Abdelhamed2019NoiseFlow} proposed a conditional flow based architecture to learn noise models. Yet, their method relies on the availability of paired data for training. Moreover, the authors employ an architecture that is specifically designed to model low-level sensor noise. In contrast, we aim to model more general degradations with no available paired training data.

\parsection{Unpaired Learning with Flows}
Whilst not for the application of learning image degradations, a few methods have investigated unpaired learning with flows. Grover \etal \cite{grover2019alignflow} trained two flow models with a shared latent space to obtain a model that adheres to exact cycle consistency. Their approach then requires an additional adversarial learning strategy based on CyCADA \cite{hoffman2017cycada}, to successfully perform domain translations.
Further, Yamaguchi \etal \cite{yamaguchi2019adaflow} proposed domain-specific normalization layers for anomaly detection. As a byproduct, their approach can perform cross-domain translations on low-resolution images, by decoding an image of one domain with the normalization layer statistics of a different domain. 
Our proposed unpaired learning approach for flows is, however, fundamentally different from these methods. We do not rely on adversarial training nor normalization layers. Instead, we introduce a shared latent space formulation that allows unpaired learning soley by minimizing the marginal negative log-likelihood.
\section{DeFlow}
\label{sec:Method}

In this paper, we strive to develop a method for learning a mapping from samples of a source domain $x \sim p_x$ to a target domain $y \sim p_y$. While there are standard supervised learning techniques for addressing this problem, paired training datasets $\{(x_i,y_i)\}_{i=1}^n$ are not available in a variety of important real-world applications. Therefore, we tackle the \emph{unpaired} learning scenario, where only unrelated sets of source $\mathcal{X} = \{x_i\}_{i=1}^n, x_i \sim p_x$ and target $\mathcal{Y} = \{y_i\}_{i=1}^m, y_i \sim p_y$ samples are available.
While we formulate a more general approach for addressing this problem, we focus on the case where $x \sim p_x$ represent non-corrupted observations, while $y \sim p_y$ are observations affected by an unknown degradation process $x \mapsto y$. In particular, we are interested in image data.

Our aim is to capture \emph{stochastic} degradation operations, which include noise and other random corruptions. The mapping $x \mapsto y$ therefore constitutes an unknown conditional distribution $p(y|x)$. The goal of this work is to learn a generative model $p(y|x;\theta)$ of this conditional distribution, without any paired samples $(x_i,y_i)$.

\subsection{Learning the Joint Distribution from Marginals}
\label{sec:1d-example}

The unpaired learning problem defined above corresponds to the task of retrieving the conditional $p(y|x)$, or equivalently, the joint distribution $p(x,y) = p(y|x)p(x)$ given only observations from the marginals $p(x)$ and $p(y)$. 
In general this is a highly ill-posed problem. However, under certain assumptions solutions can be inferred. As the most trivial case, assuming independence yields the solution $p(x,y) = p(x)p(y)$, which is not relevant since we are interested in finding correlations between $x$ and $y$. Instead, we first present a simple univariate Gaussian model, which serves as an illustrative starting point for our approach. As we will see, this example forms the simplest special case of our general DeFlow formulation.

Let us assume a 1D Gaussian random variable $x \sim p_x = \mathcal{N}(\mu_x,\sigma_x^2)$ with unknown mean $\mu_x$ and variance $\sigma_x^2$. We additionally postulate that $y = x + u$, where $u \sim p_u=\mathcal{N}(\mu_u, \sigma_u^2)$ is a Gaussian random variable that is independent of $x$.
As a sum of independent Gaussian random variables is again Gaussian, it follows that $y \sim p_y = \mathcal{N}(\mu_x + \mu_u, \sigma_x^2 + \sigma_u^2)$. Moreover, it is easy to see that $p(y|x)=\mathcal{N}(y;x+\mu_u, \sigma_u^2)$. Under these assumptions, we can estimate all unknown parameters $\theta = \{\mu_x,\sigma_x^2,\mu_u, \sigma_u^2\}$ in $p(x,y)$ by minimizing the combined negative log-likelihood of the marginal observations,
\begin{equation}
\label{eq:nll}
    L(\theta) =  - \frac{1}{n} \sum_{i=1}^n \ln p_x(x_i) - \frac{1}{m} \sum_{j=1}^m \ln p_y(y_j) \,.
\end{equation}
The derivation and resulting analytic solution is given in Appendix~\ref{supp:1d-gauss}. 
This shows that inferring the full joint distribution $p(x,y)$ given only unpaired examples is possible in this simple case. 
Next, we generalize this example using normalizing flows to achieve a highly powerful class of models capable of likelihood-based unpaired learning.

\subsection{Unpaired Learning of Conditional Flows}
\label{sec:flow}

In this section, we introduce DeFlow, a normalizing flow based formulation capable of learning flexible conditional distributions from unpaired data. Its core idea is to model the relation between $x$ and $y$ in a Gaussian latent space. We then use a deep invertible encoder-decoder network to map latent variables to the output space. 
Our model is trained end-to-end by optimizing only the marginal log-likelihoods.

We first detail the proposed latent space formulation. Our model postulates that the random variables $x \sim p_x$ and $y \sim p_y$ are related through a shared latent space. Let $z_x$ and $z_y$ denote the latent variables corresponding to $x$ and $y$ respectively. In particular, we let $z_x \sim \norm(0,I)$ follow a standard Normal distribution. The latent variable of $y$ is modeled to depend on $z_x$, but perturbed by another Gaussian random variable $u \sim p_u = \norm(\mu_u, \Sigma_u)$ such that $z_y = z_x + u$. The perturbation $u$ is independent of $x$, and therefore also of $z_x$. The mean $\mu_u$ and covariance $\Sigma_u$ of $u$ are unknown. Note that, our latent space model is the multivariate generalization of the example presented in Sec.~\ref{sec:1d-example}.

As the next step we use a powerful deep network, capable of disentangling complex patterns and correlations of \eg images to the Gaussian latent space.
In particular, we model this relation between the observations and the latent space with an invertible neural network $f_\theta$. 
Our complete model is then summarized as,
\begin{subequations}
\label{eq:model-uncond}
\begin{align}
    &x = f^{-1}_\theta(z_x) \,,\quad y = f^{-1}_\theta(z_y) = f^{-1}_\theta(z_x + u) \\
    &z_x \sim \norm(0,I) \,,\quad u \sim p_u = \norm(\mu_u, \Sigma_u) \,,\quad z_x \perp u \,.
\end{align}
\end{subequations}
Here, $\perp$ denotes stochastic independence. Note, that we can sample from the joint distribution by directly applying \eqref{eq:model-uncond}. More importantly, we can also easily sample from the conditional distribution $y_{|x} \sim p(y|x)$.
The invertibility of $f_\theta$ implies $p(y|x) = p(y|z_x)$.
From \eqref{eq:model-uncond}, we thus achieve,
\begin{equation}
\label{eq:cond-sample}
    y_{|x} = f^{-1}_\theta(f_\theta(x) + u) \sim p(y|x) \,,\quad u \sim \norm(\mu_u, \Sigma_u)\,.
\end{equation}
In words, $y_{|x}$ is obtained by first encoding $z_x = f_\theta(x)$ then sampling and adding $u$ before decoding again.

To train DeFlow with the likelihood-based objective from \eqref{eq:nll}, we employ the differentiable expressions of the marginal probability densities $p_x(x)$ and $p_y(y)$. The invertible normalizing flow $f_\theta$ allows us to apply the change of variables formula in order to achieve the expressions,
\begin{subequations}
\label{eq:marginal-density}
\begin{align}
    p_x(x) &= \big|\det D f_\theta(x)\big| \cdot \norm(f_{\theta}(x); 0, I) \\
    p_y(y) &= \big|\det D f_\theta(y)\big| \cdot  \norm(f_{\theta}(y); \mu_u, I+\Sigma_u) \,.
\end{align}
\end{subequations}
In both cases, the first factor is given by the determinant of the Jacobian $Df_\theta$ of the flow network. 
The second factors stem from the Gaussian latent space distribution of $z_x$ and $z_y$, respectively. 
For an in depth explanation of this fundamental step of normalizing flows we refer the reader to Eq.~(1) in \cite{kobyzev2020normalizing}.
It follows from \eqref{eq:cond-sample}, that $f_{\theta}(y_{|x}) = f_{\theta}(x) + u$. Therefore, we can derive the conditional density, again using  change of variables, as 
\begin{equation}
    \label{eq:cond-density}
    p(y|x) = \big|\det D f_\theta(y)\big| \cdot  \norm(f_{\theta}(y); f_{\theta}(x) + \mu_u, \Sigma_u) \,.
\end{equation}
Using \eqref{eq:marginal-density}, our model can be trained by minimizing the negative log-likelihood of the marginals \eqref{eq:nll} in the unpaired setting. Furthermore, the conditional likelihood \eqref{eq:cond-density} also enables the use of paired samples, if available. Our approach can thus operate in both the paired and unpaired setting.

It is worth noting that the 1D Gaussian example presented in Sec.~\ref{sec:1d-example} is retrieved as a special case of our model by setting the flow $f_\theta$ to the affine map
$x = f^{-1}_\theta(z) = \sigma_x z+\mu_x$. The deep flow $f_\theta$ thus generalizes our initial example beyond the Gaussian case such that complex correlations and dependencies in the data can be captured. In the case of modeling image degradations our formulation has a particularly intuitive interpretation. The degradation process $x \mapsto y$ can follow a complex and signal-dependent distribution in the image space. Our approach thus learns the bijection $f_\theta$ that maps the image to a space where this degradation can be modeled by additive Gaussian noise $u$. This is most easily seen by studying \eqref{eq:cond-sample}, which implements the stochastic degradation $x \mapsto y$ for our model. The clean data $x$ is first mapped to the latent space and then corrupted by the random Gaussian `noise' $u$. Finally, the degraded image is reconstructed with the inverted mapping $f_\theta^{-1}$.

Lastly, we note that our proposed model achieves conditioning through a very different mechanism compared to conventional conditional flows \cite{winkler2019learningConditionalFlows, srflow, Ardizzone19cINNcolor, Abdelhamed2019NoiseFlow}. These works learn a flow network that is directly conditioned on $x$ as $z = f_\theta(y;x)$. Thus, a generative model of $x$ is not learned. However, these methods rely on paired data since both $x$ and $y$ are simultaneously required to compute $z$ and its likelihood. In contrast, our approach learns the full joint distribution $p(x,y)$ and uses an unconditional flow network. The conditioning is instead performed by our latent space model \eqref{eq:model-uncond}. However, we show next that our approach can further benefit from the conventional technique of conditional flows, without sacrificing the ability of unpaired learning.

\subsection{Domain Invariant Conditioning}
\label{sec:condition}

The formulation presented in Sec.~\ref{sec:flow} requires learning the marginal distributions $p_x$ and $p_y$. For image data, this is a difficult task, requiring a large model capacity and big datasets. In this section, we therefore propose a further generalization of our formulation, which effectively circumvents the need for learning the full marginals and instead allows the network to focus on accurately learning the conditional distribution $p(y|x)$.

Our approach is based on conditioning the flow model on auxiliary information $h(x)$ or $h(y)$. Here, $h$ represents a known mapping from the observation space to a conditional variable. We use the conventional technique for creating conditional flows \cite{winkler2019learningConditionalFlows, srflow, Ardizzone19cINNcolor} by explicitly inputting $h(x)$ into the individual layers of the flow network $f_\theta$ (as detailed in Sec.~\ref{sec:architecture}). The flow is thus a function $z_x = f_\theta(x;h(x))$ that is invertible only in the first argument. Instead of the marginal distributions in \eqref{eq:marginal-density}, our approach thus models the conditional densities $p(x | h(x))$. Since $h$ is a known function, we can still learn $p(x | h(x))$ and $p(y | h(y))$ without paired data. Importantly, learning $p(x | h(x))$ is an \emph{easier} problem since information in $h(x)$ does not need modeling.

In order to ensure unpaired learning of the conditional distribution $p(y|x)$, the map $h$ must satisfy an important criterion. Namely, that $h$ only extracts \emph{domain invariant} information about the sample. Formally, this is written as,
\begin{equation}
    \label{eq:domain-invariance}
    h(x) = h(y) \,,\quad (x,y) \sim p(x,y) \,.
\end{equation}
It is easy to verify the existence of such a function $h$ by taking $h(x) = 0$ for all $x$. This choice, where $h$ carries no information about the input sample, retrieves the formulation presented in Sec.~\ref{sec:flow}. Intuitively, we wish to find a function $h$ that preserves the most information about the input, without violating the domain invariance condition \eqref{eq:domain-invariance}. Since the joint distribution $p(x,y)$ is unknown, strictly ensuring \eqref{eq:domain-invariance} is a difficult problem. In practice, however, we only need $h$ to satisfy domain invariance to the degree where it cannot be exploited by the flow network $f_\theta$. The conditioning function $h$ can thus be set empirically by gradually reducing its preserved information. We detail strategies for designing $h$ for learning image degradations in Sec.~\ref{sec:imdeg-cond}. 

\begin{figure}
\centering%
\includegraphics[width=1.\linewidth]{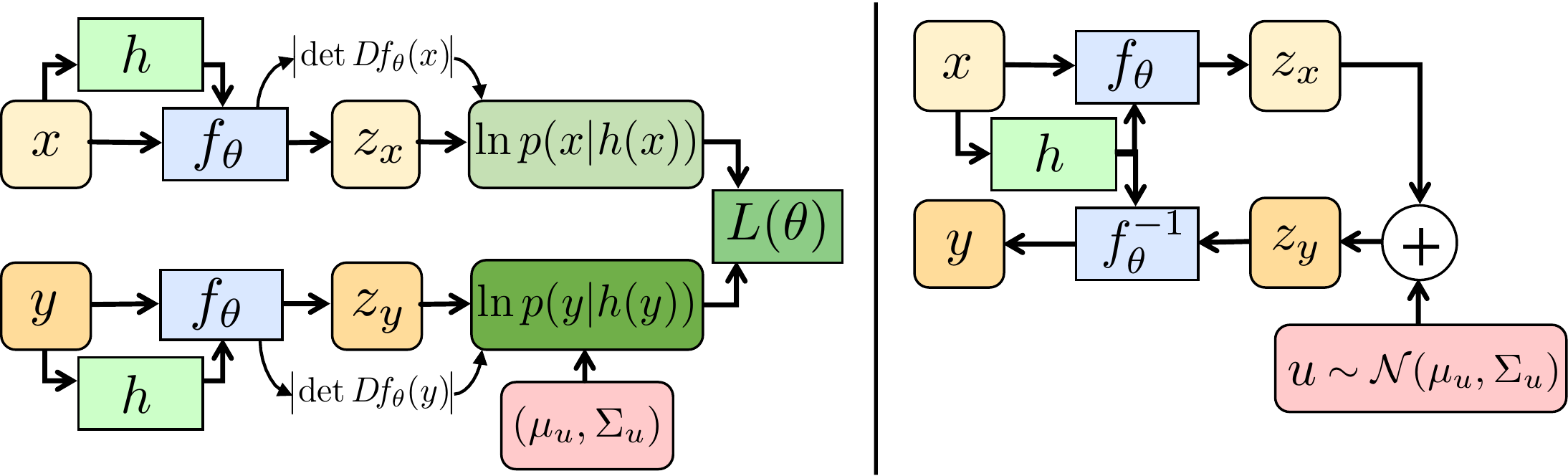}\\%
\vspace{-5pt}%
\resizebox{1.0\linewidth}{!}{%
\begin{tabular}{C{4.25cm} C{4.25cm}}%
    (a) Training & (b) Sampling\\%
\end{tabular}%
}%
\vspace{2pt}%
\caption{%
(a) DeFlow is trained to minimize the loss $L(\theta)$ in \eqref{eq:nll-h}. Unpaired inputs $x$ and $y$ are separately encoded by the flow $f_\theta$ to evaluate the NLL conditioned on $h$.
(b) We sample $y \sim p(y|x)$ using \eqref{eq:cond-sample-h} by first encoding $x$, then adding the sampled noise $u$ in the latent space and finally decoding it with the inverse flow $f_\theta^{-1}$.%
}%
\label{fig:diagram}\vspace{-4mm}
\end{figure}

The formulation in Sec.~\ref{sec:flow} is easily generalized to the case that includes the domain invariant conditioning $h$ by simply extending the flow network as $z_x = f_\theta(x; h(x))$ and $z_y = f_\theta(y; h(y))$. The training and inference stages of our resulting DeFlow formulation are visualized in Figure \ref{fig:diagram}. The model is trained by minimizing the negative log-likelihood conditioned on $h$,
\begin{equation}
\label{eq:nll-h}
    L(\theta) \!=\!  - \frac{1}{n}\! \sum_{i=1}^n \ln p(x_i|h(x_i)) - \frac{1}{m}\! \sum_{j=1}^m \ln p(y_j|h(y_j)).
\end{equation}
During inference, we sample from the conditional distribution $p(y|x)$ using,
\begin{equation}
\label{eq:cond-sample-h}
    y = f^{-1}_\theta\big(f_\theta(x;h(x)) + u;\,h(x)\big) \,,\quad u \sim \norm(\mu_u, \Sigma_u)\,.
\end{equation}
To avoid repetition, we include a detailed derivation of the generalized formulation in Appendix~\ref{supp:cond-flow}.

\section{Learning Image Degradations with DeFlow}
In this section we discuss the application of our flow-based unpaired learning formulation to the problem of generating complex image degradations. We detail the model architecture used by DeFlow and explain our approach for obtaining domain invariant conditioning in this setting.

\subsection{Model Architecture}
\label{sec:architecture}
Flow models are generally implemented as a composition of $N$ invertible layers. Let $f_\theta^n$ denote the $n$-th layer. Then the model can be expressed recursively as 
\begin{equation}
a^n = f_\theta^n(a^{n-1};h(x))
\end{equation}
where $x=a^0$, $z=a^N$ and the remaining $a^n$ represent intermediate feature maps. By the chain rule, \eqref{eq:marginal-density} gives
\begin{equation}
    p(x|h(x)) = p(z) \cdot \prod_{n=1}^N\big|\det D f_\theta^n(a^n;h(x))\big|  \\
\end{equation}
allowing for efficient log-likelihood optimization.

We parametrize the distribution $p_u$ in \eqref{eq:model-uncond} with mean $\mu_u$ the weight matrix $M$, such that $u = M \tilde{u} + \mu_u$ where $\tilde{u} \sim \norm(0,I)$ is a standard Gaussian. Consequently, the covariance is given by $\Sigma_u=MM^T$. To ensure spatial invariance, we use the same parameters $\mu_u$ and $M$ at each spatial location in the latent space. 
We initialize both $\mu_u$ and $M$ to zero, ensuring that $p(x)$ and $p(y)$ initially follow the same distribution.

Our DeFlow formulation for unsupervised conditional modeling can in principle be integrated into \emph{any} (conditional) flow architecture $f_\theta$. 
We start from the recent SRFlow~\cite{srflow} network architecture, which itself is based on the unconditional Glow~\cite{KingmaD18Glow} and RealNVP~\cite{Dinh2017RealNVP} models. We use an $L=3$ level network. Each level starts with a \emph{squeeze} operation that halves the resolution. It is followed by $K$ flow steps, each consisting of four different layers. The level ends with a \emph{split}, which removes a fraction of the activations as a latent variable. 
In our experiments we use $K=16$ flow steps, unless specified otherwise. 
Next, we give a brief description of each layer in the architecture and discuss our modifications. Please, see \cite{srflow,KingmaD18Glow} for details.

\parsection{Conditional Affine Coupling \cite{srflow}:} extends the affine coupling layer from \cite{Dinh2017RealNVP} to the conditional setting. The input feature map $a$ is split into two parts $(a_1,a_2)$ along the channel dimension. From the subset $a_1$ and the conditional $h(x)$, a scaling and bias is computed using an arbitrary neural network. These are then applied to the other subset $a_2$ providing an invertible yet flexible transformation. 

\parsection{Affine injector \cite{srflow}:} computes an individual scaling and bias for each entry of the input feature map $a$ from the conditional $h(x)$. The function computing the scaling and bias is not required to be invertible, enabling $h(x)$ to have direct influence on all channels. 

\parsection{Invertible 1x1 Convolution \cite{KingmaD18Glow}:} multiplies each spatial location with an invertible matrix. We found the LU-decomposed parametrization \cite{KingmaD18Glow} to improve the stability and conditioning of the model.

\parsection{Actnorm \cite{KingmaD18Glow}:} learns a channel-wise scaling and shift to normalize intermediate feature maps.

\parsection{Flow Step:} is the block of flow layers that is repeated throughout the network. Each flow step contains the above mentioned four layers. First, an Actnorm is applied, followed by the $1\times1$ convolution, Conditional Affine Coupling, and the Affine Injector. Note, that the last two layers are applied not only in reverse order but also in their inverted form compared to the Flow Step in SRFlow \cite{srflow}.

\parsection{Feature extraction network:} we encode the domain-invariant conditional information $h$ using the low-resolution encoder employed by SRFlow. It consists of a modified Residual-in-Residual Dense Blocks (RRDB) model \cite{wang2018esrgan}. For our experiments, we initialize it with pretrained weights provided by the authors of \cite{wang2018esrgan}. Although this network was originally intended for super-resolution, it is here employed for an entirely different task, namely to encode domain-invariant information $h$ for image degradation learning.

\subsection{Domain-Invariant Mapping $h$}
\label{sec:imdeg-cond}
The goal of our domain-invariant conditioning $h$ is to provide image information to the flow network, while hiding the domain of the input image. In our application, the domain invariance \eqref{eq:domain-invariance} implies that the mapping $h$ needs to remove information that could reveal whether input is a clean $x$ or a degraded $y$ image. On the other hand, we want to preserve information about the underlying image content to simplify learning. We accomplish this by utilizing some prior assumptions that are valid for most stochastic degradations. Namely, that they mostly affect the high frequencies in the image, while preserving the low frequencies. 

We construct $h$ by down-sampling the image to a sufficient extent to remove the visible impact of the degradations. We found it beneficial to also add a small amount of noise to the resulting image to hide remaining traces of the original degradation. The domain invariant mapping is thus constructed as $h(x)=d_\downarrow(x)+n,\, n\sim\norm(0,\sigma^2)$, where $d_\downarrow(x)$ denotes bicubic downsampling. Note that this operation is only performed to extract a domain-invariant representation, and is not related to the degradation $x \mapsto y$ learned by DeFlow. The purpose of $h$ is to \emph{remove} the original degradation, while preserving image content.

\begin{figure*}[t]
    \centering%
    \newcommand{\size}{0.1315}%
    \includegraphics[width=\size\linewidth]{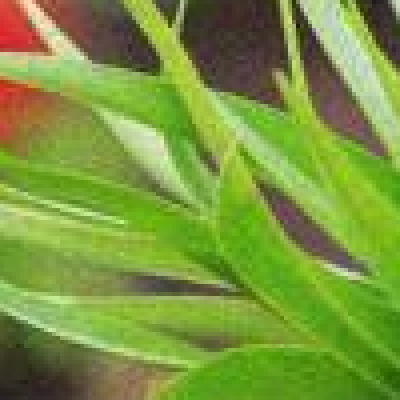}~~~~%
    \includegraphics[width=\size\linewidth]{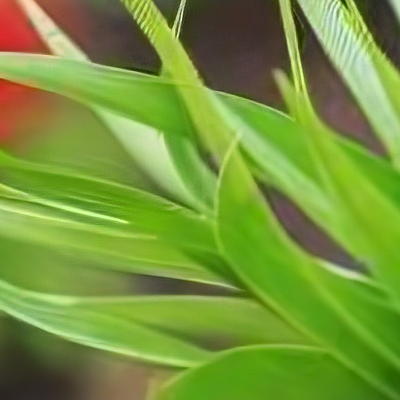}~~~~%
    \includegraphics[width=\size\linewidth]{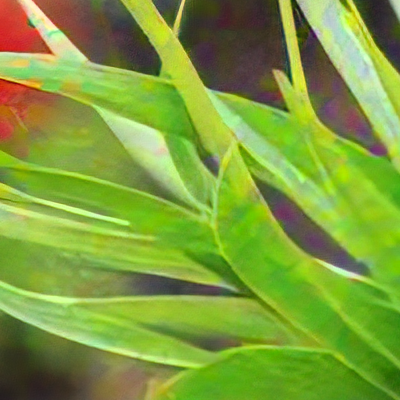}~~~~%
    \includegraphics[width=\size\linewidth]{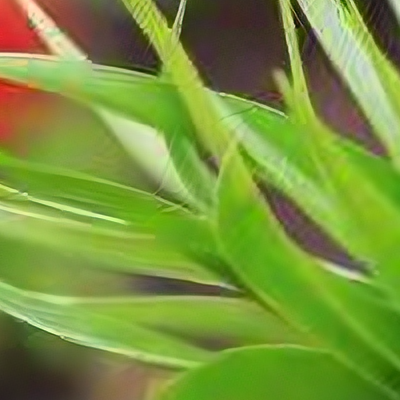}~~~~%
    \includegraphics[width=\size\linewidth]{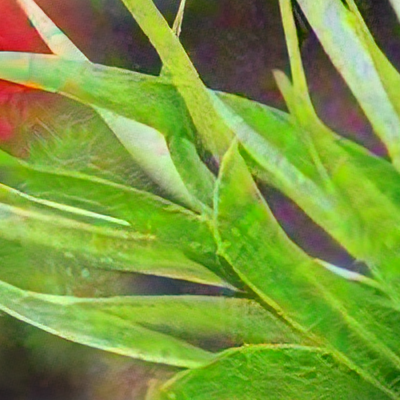}~~~~%
    \includegraphics[width=\size\linewidth]{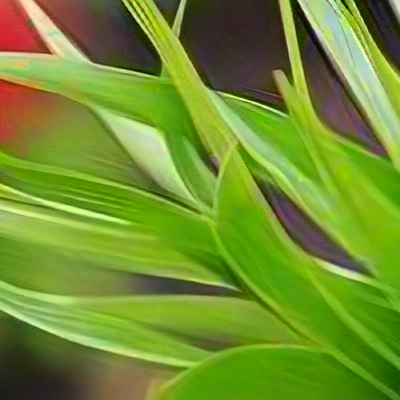}~~~~%
    \includegraphics[width=\size\linewidth]{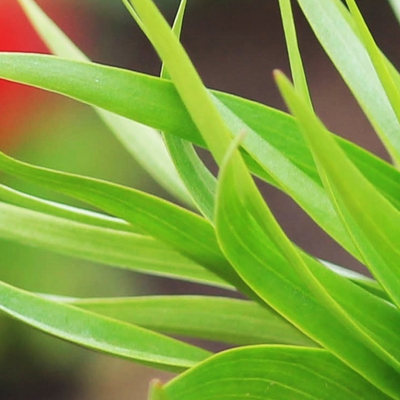}%
    \vspace{-6pt}
    \resizebox{1.02\linewidth}{!}{%
    \centering
    \begin{tabular}{C{4cm} C{4cm} C{4cm} C{4cm} C{4cm} C{4cm} C{4cm}}
        LR  & White Noise $\sigma\!=\!0.04$ & DASR$^\dagger$ \cite{wei2020unsupervised}& ~Frequency Separation$^\dagger$ \cite{Fritsche19Frequency} & ~~Impressionism$^\dagger$ \cite{Ji_2020_CVPR_Workshops} & ~\textbf{DeFlow} (ours) & GT
    \end{tabular}
    }

    \includegraphics[width=\size\linewidth]{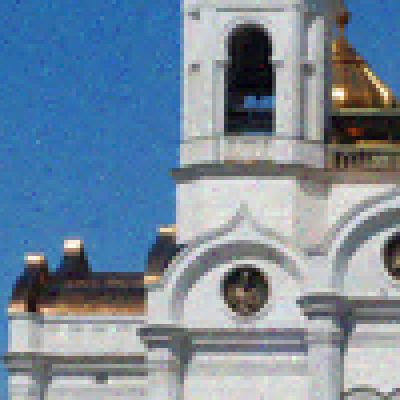}~~~~%
    \includegraphics[width=\size\linewidth]{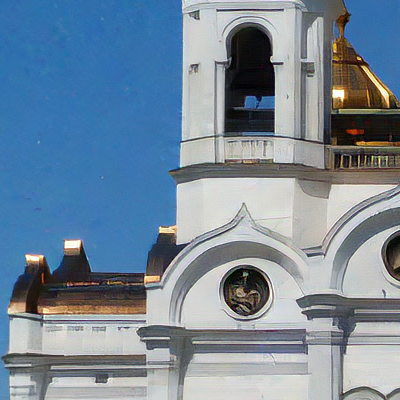}~~~~%
    \includegraphics[width=\size\linewidth]{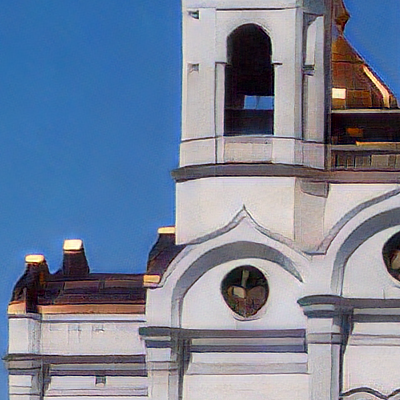}~~~~%
    \includegraphics[width=\size\linewidth]{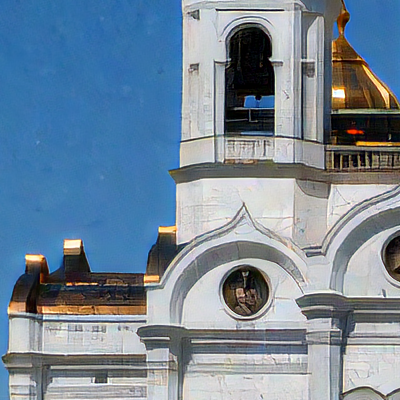}~~~~%
    \includegraphics[width=\size\linewidth]{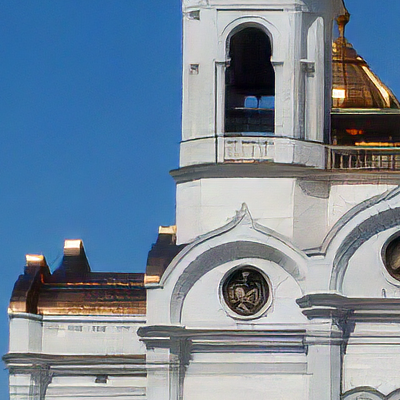}~~~~%
    \includegraphics[width=\size\linewidth]{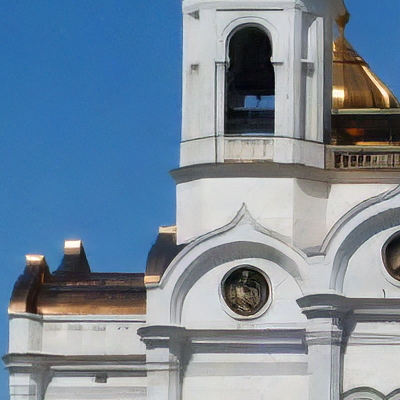}~~~~%
    \includegraphics[width=\size\linewidth]{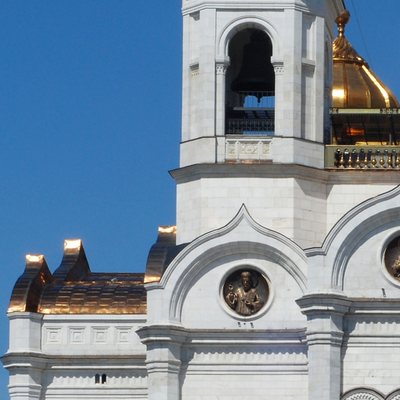}%
    \vspace{-6pt}
    \resizebox{1.02\linewidth}{!}{%
    \centering
    \begin{tabular}{ C{4cm} C{4cm} C{4cm} C{4cm} C{4cm} C{4cm} C{4cm}}
        LR  & White Noise $\sigma\!\sim\!\mathcal{U}(0,0.06)$~~ & CycleGAN$^\dagger$ \cite{AIM2019RWSRchallenge} & ~Frequency Separation$^\dagger$ \cite{Fritsche19Frequency} & ~~Impressionism$^\dagger$ \cite{Ji_2020_CVPR_Workshops} & \textbf{DeFlow} (ours) & GT
    \end{tabular}
    }%

    \includegraphics[width=\size\linewidth]{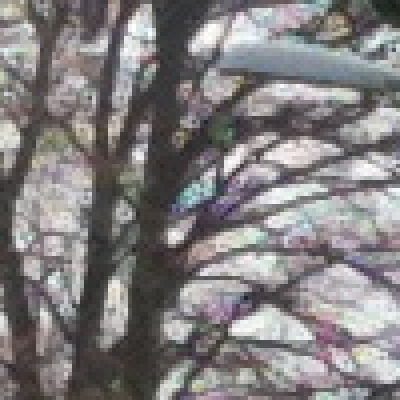}~~~~%
    \includegraphics[width=\size\linewidth]{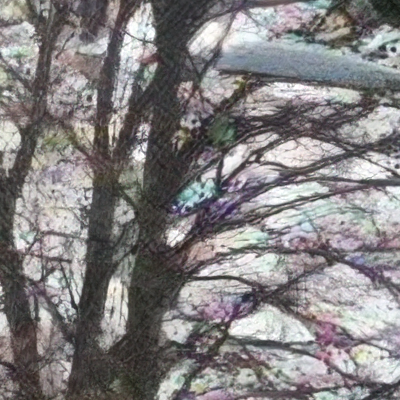}~~~~%
    \includegraphics[width=\size\linewidth]{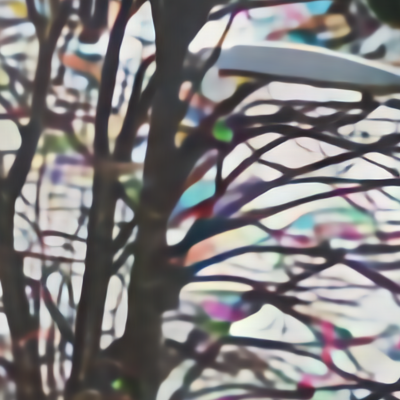}~~~~%
    \includegraphics[width=\size\linewidth]{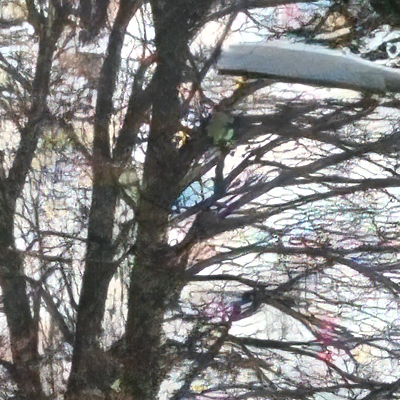}~~~~%
    \includegraphics[width=\size\linewidth]{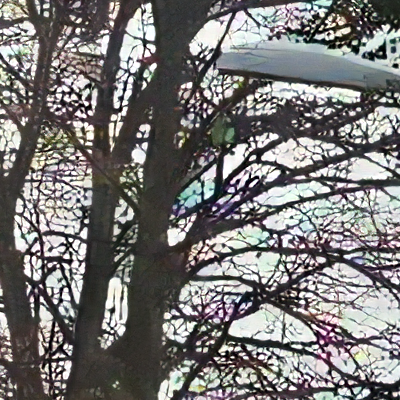}~~~~%
    \includegraphics[width=\size\linewidth]{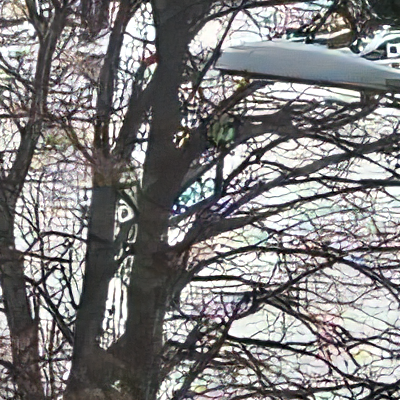}~~~~%
    \includegraphics[width=\size\linewidth]{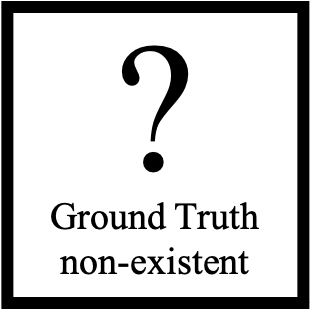}%

    \vspace{-6pt}
    \resizebox{1.02\linewidth}{!}{%
    \centering
    \begin{tabular}{ C{4cm} C{4cm} C{4cm} C{4cm} C{4cm} C{4cm} C{4cm}}
        LR  & ~~No Degradation \cite{wang2018esrgan} & CycleGAN \cite{AIM2019RWSRchallenge} & ~Frequency Separation$^\dagger$ \cite{Fritsche19Frequency} & ~~Impressionism \cite{Ji_2020_CVPR_Workshops} & \textbf{DeFlow} (ours) & GT
    \end{tabular}
    }%

    \caption{Super-resolved images from the AIM-RWSR (top), NTIRE-RWSR (mid) and DPED-RWSR (bottom) datasets. Top-5 methods are shown based on LPIPS score for the synthetic datasets and the visual judgement of the authors for the DPED-RWSR dataset.}%
    \label{fig:sr_visualization}%
    \vspace{-9pt}
\end{figure*}

\section{Experiments and Results}
We validate the degradations learned by DeFlow by applying them to the problem of real-world super-resolution (RWSR). Here, the task is to train a joint image restoration and super-resolution model without paired data that is able to translate degraded low-resolution images to high-quality and high-resolution images. 
In particular, we employ DeFlow to learn the underlying degradation model and use it to generate paired training data for a supervised super-resolution model. Experiments are performed on three recent benchmarks designed for this setting. Detailed results with more visual examples are shown in Appendix~\ref{supp:DeFlow-results}.

\subsection{Datasets}
\label{sec:datasets}
\parsection{AIM-RWSR:} Track 2 of the AIM 2019 RWSR challenge \cite{AIM2019RWSRchallenge} provides a dataset consisting of a source and a target domain. The former contains synthetically degraded images from the Flickr2k dataset \cite{timofte2017ntire} that feature some combination of noise and compression, while the latter contains the high-quality non-degraded images of the DIV2k dataset~\cite{div2k}. The task is to $4\times$ super-resolve images from the source domain to high-quality images as featured in the target domain. Since the degradations were generated synthetically, there exists a validation set of 100 paired degraded low-resolution and high-quality ground-truth images, allowing the use of reference-based evaluation metrics. 

\parsection{NTIRE-RWSR:} Track 1 of the NTIRE 2020 RWSR challenge \cite{ntire2020RWSR} follows the same setting as AIM-RWSR. 
However, it features a completely different type of degradation, namely highly correlated high-frequency noise. 
As before, a validation set exists enabling a reference-based evaluation.

\parsection{DPED-RWSR:} Differing from the other two datasets, the source domain of Track 2 of the NTIRE 2020 RWSR challenge consists of real low-quality smartphone photos that are to be jointly restored and super-resolved. A high-quality target domain dataset is also provided. The source domain stems from the iPhone3 images of the DPED dataset \cite{ignatov2017dslr}, while the target domain corresponds to the DIV2k~\cite{div2k} training set. Because reference images do not exist evaluation is restricted to no-reference metrics and visual inspection.

\subsection{Evaluation Metrics}
For the synthetic datasets, we report the peak signal-to-noise ratio (PSNR) and the structural similarity index (SSIM) \cite{WangBSS04SSIM}. In addition we compute the Learned Perceptual Image Patch Similarity (LPIPS) \cite{zhang2018unreasonable} metric, a reference-based image quality metric based on feature distances in CNNs. As LPIPS has been shown to correlate well with human perceived image quality, we consider it the most important metric for the RWSR task.
For the DPED-RWSR we report the NIQE \cite{MittalSB13NIQE}, BRISQUE \cite{mittal2011brisque} and PIQE \cite{NDBCM15piqe} no-reference metrics.
We also conduct a user study comparing the best models with our DeFlow method. For each compared method, we show participants randomly selected crops super-resolved by both DeFlow and the compared method. Users are then asked to select the more realistic image. We report user preference as the percentage of images where the compared method was preferred over DeFlow.
A User Preference $<\!\!50\%$ indicates that DeFlow obtains `better' images than the comparison method. More details on the user study are provided in Appendix ~\ref{supp:user-study}.

\subsection{Baselines and other Methods}
We compare DeFlow against Impressionism \cite{Ji_2020_CVPR_Workshops} the winner of the NTIRE 2020 RWSR challenge \cite{ntire2020RWSR} and Frequency Separation \cite{Fritsche19Frequency} the winner of the AIM 2019 RWSR challenge \cite{AIM2019RWSRchallenge}. Further, we compare with the very recent DASR \cite{wei2020unsupervised} and the CycleGan based method introduced in \cite{lugmayrICCVW2019}. All aforementioned methods apply the same two-stage approach, where first a degradation model is learned  to generate synthetic training data that is then used to train a supervised ESRGAN \cite{wang2018esrgan} based super-resolution model. 
We also validate against simple baselines. 
Our \emph{No Degradation} baseline is trained without any degradation model. The \emph{White Noise} model adds zero mean Gaussian noise to the low-resolution patches during training. Here, we tested two variants, either fixing the standard deviation $\sigma$ or sampling the standard deviation of the noise added to each image patch uniformly from $\mathcal{U}(0,\sigma_{max})$.
For each dataset we tuned both variants with different choices of $\sigma$ and $\sigma_{max}$, respectively, and only report the model that obtained the best LPIPS score.

\subsection{Training Details} 
\label{sec:DeFlow_train_settings}
We train all DeFlow models for 100k iterations using the Adam~\cite{KingmaB14Adam} optimizer. The initial learning rate is set to $5\cdot 10^{-5}$ on the synthetic datasets and to $5\cdot 10^{-6}$ on the DPED-RWSR dataset and is halved at 50k, 75k, 90k and 95k iterations.
We use a batch size of 8 with random crops of size $160 \times 160$ on the AIM-RWSR and NTIRE-RWSR dataset. On DPED-RWSR we obtained better performance with a patch size of $80 \times 80$ and a batch size of $48$. Batches are sampled randomly such that images of both domains are drawn equally often.
Random flips are used as a data augmentation.
We use $64$ hidden channels in the affine injector layer for NTIRE-RWSR and DPED-RWSR and $128$ on AIM-RWSR.
Similar to \cite{KingmaD18Glow,srflow}, we apply a 5-bit de-quantization by adding uniform noise to the input of the flow model. 
We train the DeFlow models using the $4\times$ bicubic downsampled clean domain as $\mathcal{X}$ and the noisy domain as $\mathcal{Y}$. 
Given the large domain gap between the source and target images in DPED-RWSR we do not use the target images and instead use $4\times$ and $8\times$ bicubic downsampled noisy images as the clean domain $\mathcal{X}$. 
For DPED-RWSR we further follow the approach of \cite{Ji_2020_CVPR_Workshops} and estimate blur kernels of the degraded domain using KernelGAN \cite{bell2019blindKernelGAN}. These are then applied to any data from the clean domain, \ie on the clean training data and before degrading images.
On AIM-RWSR we normalize $\mathcal{X}$ and $\mathcal{Y}$ to the same channel-wise means and standard deviations. Degraded images are then de-normalized before employing them as training data for the super-resolution model.
For the conditional $h(x)$ we used $\sigma=0.03$ in conjunction with $4\times$ bicubic downsampling on NTIRE-RWSR and DPED-RWSR and $8\times$ bicubic downsampling on AIM-RWSR.


\begin{table}
\centering
\resizebox{\columnwidth}{!}{
\begin{tabular}{lccc|c}
\toprule
  &   PSNR$\uparrow$ & SSIM$\uparrow$ & LPIPS$\downarrow$ & User Pref.\ \\
\midrule

CycleGan \cite{AIM2019RWSRchallenge}                        &  21.19            &  0.53             &  0.476            &   - \\
Frequency Separation \cite{Fritsche19Frequency}             &  21.00            &  0.50             &  0.403            &   \textcolor{Red}{38.89\%} \\
DASR \cite{wei2020unsupervised}                             &  21.79            &  0.58             &  0.346            &   \textcolor{Red}{35.74\%} \\

\midrule
No Degradation                     &  21.82 &  0.56 &  0.514 &   - \\
Impressionism$^\dagger$ \cite{Ji_2020_CVPR_Workshops}       & \textbf{22.54}    &  0.63             &  0.420            &   \textcolor{Red}{27.58\%} \\
White Noise $\sigma\!=\!0.04$                               &  22.43            &  \textbf{0.65}    &  0.406            &   \textcolor{Red}{30.00\%} \\
Frequency Separation$^\dagger$ \cite{Fritsche19Frequency}   &  20.47            &  0.52             &  0.394            &   \textcolor{Red}{39.37\%} \\
DASR$^\dagger$ \cite{wei2020unsupervised}                   &  21.16            &  0.57             &  0.370            &   \textcolor{Red}{40.26\%} \\
\textbf{DeFlow} (ours)                                      &  22.25            &  0.62             &  \textbf{0.349}   &   reference \\

\bottomrule
\end{tabular}}\vspace{1mm}
\caption{AIM-RWSR results: Methods in the bottom segment use the same SR pipeline.
User preferences are \textcolor{Green}{green} if the method and \textcolor{Red}{red} if DeFlow was preferred by the majority. \textcolor{Orange}{Orange} indicates a result within the 95\% confidence interval.}\vspace{-2mm}
\label{table:aim}
\end{table}

\begin{table}
\centering
\resizebox{\columnwidth}{!}{
\begin{tabular}{lccc|c}
\toprule
{} &    PSNR$\uparrow$ & SSIM$\uparrow$ & LPIPS$\downarrow$ &  User Pref.\ \\
\midrule
CycleGan\cite{AIM2019RWSRchallenge}                         &  24.75            &  0.70             &  0.417            & \textcolor{Red}{35.78\%} \\
Impressionism \cite{Ji_2020_CVPR_Workshops}                 &  24.77            &  0.67             &  0.227            & \textcolor{Green}{54.11\%} \\
\midrule
No Degradation                                              &  20.59            &  0.34             &  0.659            & - \\
Frequency Separation$^\dagger$ \cite{Fritsche19Frequency}~~ &  23.04            &  0.59             &  0.332            & \textcolor{Red}{46.17\%} \\ 
CycleGan$^\dagger$ \cite{AIM2019RWSRchallenge}              &  22.62            &  0.60             &  0.314            & \textcolor{Red}{44.72\%} \\
White Noise $\sigma\!\sim\!\mathcal{U}(0,0.06)$             &  25.47            &  \textbf{0.71}    &  0.237            & \textcolor{Green}{53.28\%} \\
Impressionism$^\dagger$ \cite{Ji_2020_CVPR_Workshops}       &  25.03            &  0.70             &  0.226            & \textcolor{Green}{56.44\%} \\
\textbf{DeFlow} (ours)                                      &  \textbf{25.87}   &  \textbf{0.71}    &  \textbf{0.218}   & reference \\

\bottomrule
\end{tabular}}\vspace{1mm}
\caption{NTIRE-RWSR results: see caption in Tab.~\ref{table:aim}.}\vspace{-2mm}
\label{table:ntire}
\end{table}

\begin{table}[t]
\centering
\resizebox{\columnwidth}{!}{%
\begin{tabular}{lccc|c}
\toprule
{} &  NIQE $\downarrow$ & BRISQUE$\downarrow$ & PIQE$\downarrow$ & User Pref.\ \\
\midrule
CycleGAN \cite{AIM2019RWSRchallenge}                        &  5.47             &    49.19          &  86.83            &   -           \\
Frequency Separation \cite{Fritsche19Frequency}             &  3.27             &    22.73          &  11.88            &   \textcolor{Red}{36.88\%}     \\
Impressionism \cite{Ji_2020_CVPR_Workshops}                 &  4.12             &    23.24          &  14.09            &   \textcolor{Green}{54.13\%}     \\
\midrule
No Degradation                                              &  3.55             &    24.56          &   \textbf{8.01}   &   -           \\
KernelGAN$^\dagger$  \cite{bell2019blindKernelGAN}          &  6.37             &    42.74          &  30.32            &   -           \\
Frequency Separation$^\dagger$ \cite{Fritsche19Frequency}   &  \textbf{3.39}             &    25.40          &  11.22            &   \textcolor{Red}{37.48\%}     \\
Impressionism$^\dagger$ \cite{Ji_2020_CVPR_Workshops}       &  3.85             &    21.49          &  12.84            &   \textcolor{Orange}{50.72\%}     \\
\textbf{DeFlow} (ours)                                      &  3.42             &    \textbf{21.13} &  15.84            &   reference   \\

\bottomrule
\end{tabular}}\vspace{1mm}
\caption{DPED-RWSR results: see caption in Tab.~\ref{table:aim}.}\vspace{-2mm}
\label{table:dped}
\end{table}

\subsection{Super-Resolution Model}
\label{sec:ESRGAN_train_settings}
To fairly compare with existing approaches, we an ESRGAN \cite{wang2018esrgan} as the super-resolution model. Specifically, we employ the training code provided by the authors of Impressionism \cite{Ji_2020_CVPR_Workshops} that trains a standard ESRGAN for 60k iterations. For AIM-RWSR and NTIRE-RWSR the standard VGG discriminator is used while on DPED-RWSR a patch discriminator is applied. As in \cite{Ji_2020_CVPR_Workshops}, we use the $2\times$ down-sampled smartphone images of the DPED-RWSR dataset as clean images and do not use the provided high-quality data. Unlike \cite{Ji_2020_CVPR_Workshops} however, we do not use any downsampled noisy images as additional clean training data. 
We evaluate the trained models after 10k, 20k, 40k and 60k iterations and report the model with the best LPIPS on the validation set. For DPED-RWSR we simply choose the final model.
To better isolate the impact of the learned degradations, we further report the performance of other methods when using their degradation pipeline with our super-resolution model. We mark these models with the $\dagger$ symbol.

\subsection{Comparison with State-of-the-Art}
First, we discuss the results on the AIM-RWSR dataset shown in Tab.~\ref{table:aim}.
The GAN-based Frequency Separation approach \cite{Fritsche19Frequency}, the winner of this dataset's challenge, obtains an LPIPS similar to the White Noise baseline. DASR \cite{wei2020unsupervised} obtains a highly competitive LPIPS, yet it is strongly outperformed by DeFlow in our user study.
In fact, as shown in Fig.~\ref{fig:sr_visualization}, DASR generates strong artifacts. This can be explained by overfitting, as DASR directly optimizes for LPIPS during training. When using the degradation model of DASR in conjunction with our super-resolution pipeline the resulting model DASR$^\dagger$ performs slightly better in the user study while obtaining an LPIPS score of $0.370$ compared to DeFlow's $0.349$. 
Notably, DeFlow outperforms all previous methods by a large margin in the user study. It also obtains a higher PSNR and SSIM than all methods with learned, but GAN based degradation models.

On the NTIRE-RWSR dataset (see Tab.~\ref{table:ntire}) DeFlow obtains the best scores among all reference metrics, making it the only model that consistently outperforms the White Noise baseline. In the user study DeFlow is also preferred to all learned degradation models.
Yet, the user study indicates better quality from the hand-crafted degradation models, namely Impressionism and the White Noise baseline, compared to the learned approach of DeFlow. However, as shown in the second row of Fig.~\ref{fig:sr_visualization}, the White Noise baseline generates highly visible artifacts in smooth regions, \eg sky, whereas DeFlow removes all noise from these areas. 

Lastly, we compare the results on the DPED-RWSR dataset in Tab.~\ref{table:dped}. 
Similar to \cite{ntire2020RWSR}, we find that the no-reference metrics do not correlate well with the perceived quality of the images.
As shown in Fig.~\ref{fig:sr_visualization}, DeFlow obtains sharp images with pleasing details clearly outperforming all other learned approaches. Compared to Impressionism \cite{Ji_2020_CVPR_Workshops}, we find that our method produces fewer artifacts and does not over-smooth textures. However, we notice that our images retain more noise and are sometimes less sharp.
This is supported by the user study where DeFlow significantly outperforms the Frequency Separation method \cite{Fritsche19Frequency}, while being head-to-head with Impressionism$^\dagger$ \cite{Ji_2020_CVPR_Workshops}. 

Overall, DeFlow is the only method with consistently good performance across all three datasets, whereas the handcrafted approaches obtain the worst performance on the AIM-RWSR dataset and the other learned approaches are struggling to create artifact-free yet detailed images on the NTIRE-RWSR dataset. 
It is also noteworthy that CycleGAN \cite{AIM2019RWSRchallenge}, despite its immense popularity for unpaired learning, does not perform well on any of these datasets. This can be partly explained by the weak cycle consistency constraint and the use of a deterministic generator.

\subsection{Ablation Study}
In this section, we analyze DeFlow through an ablation study. 
We train a variety of models on the AIM-RWSR dataset and evaluate their downstream super-resolution performance. These models deviate only in the choice of a single hyper-parameter with all other training settings remaining as described in \ref{sec:DeFlow_train_settings}. 
In particular, we scrutinize on three core segments: the depth of the model, the choice of conditioning $h(x)$, and the method of learning the domain shift. For each segment we show the results of this study in a separate section of Tab.~\ref{table:ablation}. 

\begin{table}[t]
\centering%
\resizebox{\columnwidth}{!}{
    \begin{tabular}{lccc}
    \toprule
    {} &   PSNR$\uparrow$ & SSIM$\uparrow$ & LPIPS$\downarrow$ \\
    \midrule
    $K=4$ Flow Steps            &  22.18 &  0.61 &  0.362 \\
    $K=8$ Flow Steps            &  22.20 &  0.63 &  0.355 \\
    \textbf{$K=16$ Flow Steps}  &  22.25 &  0.62 &  0.349 \\
    \midrule
    $4\times$ downsampling in $h(x)$   &  22.44 &  0.61 &  0.429 \\
    \textbf{$8\times$ downsampling in $h(x)$}   &  22.25 &  0.62 &  0.349 \\
    $16\times$ downsampling in $h(x)$  &  21.52 &  0.61 &  0.352 \\
    No Conditional $h(x)=0$     &  18.33 &  0.52 &  0.412 \\
    \midrule
    Non-learned Shift             &  22.04 &  0.62 &  0.405 \\
    Learned uncorrelated Shift          &  22.04 &  0.61 &  0.359 \\
    \textbf{Learned correlated Shift}~~~~~~~~~~~~~    &  ~~22.25~~ &  ~~0.62~~ &  ~~0.349~~ \\

    \bottomrule
    \end{tabular}%
}%
\vspace{1mm}%
\caption{Ablation study of DeFlow on the AIM-RWSR dataset. The final setting of the DeFlow model for AIM-RWSR are in \textbf{bold}.}%
\label{table:ablation}%
\vspace{-4mm}
\end{table}%

\parsection{Network depth (Tab.~\ref{table:ablation}, top):} Increasing the number of Flow Steps $K$ improves performance, showing that indeed powerful networks help to learn the complex degradations.

\parsection{Conditioning (Tab.~\ref{table:ablation}, middle):} Next we analyze the impact of the domain invariant conditioning $h(x)$ (Sec.~\ref{sec:condition}). Using $4\times$ downsampling in the conditional yields noticeable worse performance compared to larger factors. We conclude that larger downsampling factors are required to ensure the domain invariance of $h(x)$.
Notably, $16\times$ downsampling yields only a slight performance reduction compared to $8\times$ downsampling. In contrast, no conditional information at all \ie $h(x)=0$ leads to a significantly worse performance where the translated images exhibits strong color shifts and blur. This highlights the importance of the conditional and shows that even little auxiliary information yields drastic performance improvements. 

\parsection{Learned shift (Tab.~\ref{table:ablation}, bottom):} Last, we investigate our latent space formulation. We first restrict the added noise $u \sim p_u$ to be uncorrelated across the channels by constraining $\Sigma_u$ to a diagonal covariance matrix. We notice a negative impact on performance. This demonstrates the effectiveness of our more general Gaussian latent space model.
Further, we validate our choice of using domain dependent base distributions. We train a DeFlow model with a standard normal Gaussian as the base distribution for both domains (\ie setting $u\!=\!0$ in \eqref{eq:model-uncond}). We then infer the domain shift after training by computing the channel-wise mean and covariance matrix in the latent space for each domain. The resulting empirical distributions of both domains become very similar and the inferred shift does no longer model the domain shift faithfully. This results in a substantially worse performance in the down-stream task and further shows the potential of our unpaired learning formulation.
\section{Conclusion}
We propose DeFlow, a method for learning conditional flow networks with unpaired data. Through a constrained latent space formulation, DeFlow learns the conditional distribution by minimizing the marginal negative log-likelihoods. We further generalize our approach by conditioning on domain invariant information.  We apply DeFlow to the unsupervised learning of complex image degradations, where the resulting model is used for generating training data for the downstream task of real-world super-resolution. Our approach achieves state-of-the-art results on three challenging datasets.

\parsection{Acknowledgements}
This work was partly supported by the ETH Z\"urich Fund (OK), a Huawei Technologies Oy (Finland) project, an Amazon AWS grant, a Microsoft Azure grant, and a Nvidia hardware grant.

{\small
\bibliographystyle{ieee_fullname}
\bibliography{references}
}

\clearpage
\clearpage
\newpage
\begin{center}
\Large\textbf{Appendix}
\end{center}
\appendix

In Sec.~\ref{supp:1d-gauss} of this appendix, we first derive the closed-form solution of the 1D Gaussian example from Sec.~\ref{sec:1d-example}
. We then go on in Sec.~\ref{supp:restricted-base} and show that restricting $p_x$ to a standard normal distribution is absorbed by a single affine layer in the deep flow model.
Next, we provide a derivation to the DeFlow method with domain invariant conditioning in Sec.~\ref{supp:cond-flow}.
We then show in Sec.~\ref{supp:DeFlow-results} that degradations generated by DeFlow are stochastic and can be sampled at varying strengths. Further, we provide a visual comparison of the degradations and more example images of the downstream real-world super-resolution (RWSR) performance in Sec.~\ref{supp:visual-comp}. Lastly, we give insight into the set-up of the conducted user study in Sec.~\ref{supp:user-study}.

\section{Closed-Form Solution for the \mbox{1D} Gaussian Example} \label{supp:1d-gauss}
Here we present a detailed derivation for the closed-form solution to the 1-dimensional Gaussian example from Sec.~\ref{sec:1d-example}. 
To recall, we are given two datasets $\mathcal{X}=\{x_i\}_{i=1}^N$ and $\mathcal{Y}=\{y_i\}_{i=1}^M$. We know that $x \in \mathcal{X}$ are i.i.d.\ samples from $p_x = \mathcal{N}(\mu_x,\sigma_x^2)$. Further, we know that $y=x+u \in \mathcal{Y}$ are i.i.d.\ samples from $x\sim p_x$ with additive independent Gaussian noise $u\sim p_u=\mathcal{N}(\mu_u, \sigma_u^2)$.

The task is to find the parameters $\theta^* = \{\mu_x, \sigma_x^2, \mu_u, \sigma_u^2\}$ that jointly maximize the marginal likelihoods  $p_x(\mathcal{X})$ and $p_y(\mathcal{Y})$.

Proceeding as usual, we apply the i.i.d.\ property and minimize the negative log-likelihood \wrt $\theta$,
\begin{align}
    \min l(\theta) =& -\frac{1}{N}\sum_{i=1}^N \frac{(x_i-\mu_x)^2}{\sigma_x^2}+\frac{\ln \sigma_x^2}{2} \nonumber\\
    &-\frac{1}{M}\sum_{i=1}^N \frac{(y_i-\mu_x-\mu_u)^2}{\sigma_x^2+\sigma_u^2}+\frac{\ln (\sigma_x^2 + \sigma_u^2)}{2} \nonumber\\
    \text{subject to} \quad & \sigma_x \ge 0 \,,\; \sigma_u \ge 0 \,.
\end{align}
To ensure the estimated variances are non-negative, \ie $\sigma_x \ge 0$ and  $\sigma_u \ge 0$, we introduce the Lagrange multipliers $ \lambda_x $ and $ \lambda_u $ and have,
\begin{equation}
    \hat{l}(\theta) = l(\theta) - \lambda_x \sigma_x^2 - \lambda_u \sigma_u^2 \,.
\end{equation}
By the Karush–Kuhn–Tucker theorem, $\theta^*$ is a optimal solution to $l(\theta)$ if $\frac{\partial \hat{l}(\theta^*)}{\partial \theta}=0$ while $\lambda_x \ge 0$, $\lambda_u \ge 0$, $\lambda_x \sigma_x^2 = 0$ and $\lambda_u \sigma_u^2 = 0$ hold.

Next, we take partial derivatives of $\hat{l}(\theta)$ \wrt the individual parameters and set them to $0$ to obtain the optimal estimates. First, we differentiate \wrt the means $\mu_x$ and $\mu_u$, and obtain 
\begin{align}
    \label{eq:d_mu_u}
    \frac{\partial \hat{l}(\theta)}{\partial \mu_u} &= \frac{1}{M} \frac{1}{\sigma_x^2+\sigma_u^2} \sum_{y\in \mathcal{Y}} (y - (\mu_x + \mu_u)) \doteq 0\\
    \frac{\partial \hat{l}(\theta)}{\partial \mu_x} &= \frac{1}{N} \frac{1}{\sigma_x^2} \sum_{x\in \mathcal{X}} (x - \mu_x) \nonumber\\ 
    &+ \frac{1}{M} \frac{1}{\sigma_x^2+\sigma_u^2} \sum_{y\in \mathcal{Y}} (y - (\mu_x + \mu_u)) \nonumber\\
    &\stackrel{(\ref{eq:d_mu_u})}{=} \frac{1}{N} \frac{1}{\sigma_x^2} \sum_{x\in \mathcal{X}} (x - \mu_x) \doteq 0 \,.
    \label{eq:d_mu_x}
\end{align}
It directly follows, that the optimal estimates of $\mu_x$ and $\mu_u$ can be written as the empirical means $\hat{\mu}_x$ and $\hat{\mu}_y$,
\begin{align}
    \label{eq:mu_x}
    \mu_x &= \hat{\mu}_x = \frac{1}{N} \sum_{x\in \mathcal{X}} x \\
    \mu_u &= \hat{\mu}_y - \hat{\mu}_x \,,\quad \hat{\mu}_y  =  \frac{1}{M} \sum_{y\in \mathcal{Y}} y \,. 
\end{align}

Now we turn to the estimation of the variances. We first obtain the following partial derivatives,
\begin{align}
    \frac{\partial \hat{l}(\theta)}{\partial \sigma_u^2} =& \frac{1}{2(\sigma_x^2+\sigma_u^2)}
    -\frac{1}{2M}\sum_{y\in \mathcal{Y}} \frac{(y-\hat{\mu_y})^2}{(\sigma_x^2+\sigma_u^2)^2}
    -\lambda_u \\
    \frac{\partial \hat{l}(\theta)}{\partial \sigma_x^2} =& 
    \frac{1}{2\sigma_x^2} 
    -\frac{1}{2N} \sum_{x \in \mathcal{X}} \frac{(x-\mu_x)^2}{\sigma_x^4} \nonumber\\
    +& \frac{1}{2(\sigma_x^2+\sigma_u^2)}
    -\frac{1}{2M}\sum_{y\in \mathcal{Y}} \frac{(y-\hat{\mu_y})^2}{(\sigma_x^2+\sigma_u^2)^2}
    -\lambda_x \,.
\end{align}
Setting $\frac{\partial \hat{l}(\theta)}{\partial \sigma_u^2}$ to $0$ and using the complementary slackness condition that $\lambda_u \sigma_u = 0$ must hold at the minimum we obtain,
\begin{align}
    &\frac{\partial \hat{l}(\theta)}{\partial \sigma_u^2} \doteq 0\\
   \iff& -(\sigma_x^2+\sigma_u^2) + 2\lambda_u(\sigma_x^2+\sigma_u^2)^2 + \hat{\sigma}_y^2 = 0\\
    \iff&  2\lambda_u \sigma_x^4  - \sigma_x^2 - \sigma_u^2 + \hat{\sigma}_y^2 = 0\\
    \iff& \sigma_u^2 = 2\lambda_u \sigma_x^4  - \sigma_x^2 + \hat{\sigma}_y^2 \,.
    \label{eq:var_u_deriv}
\end{align}
where $\hat{\sigma}_y = \frac{1}{M}\sum_{y\in \mathcal{Y}} (y-\hat{\mu_y})^2$ is used as short-hand notation for the empirical variance of $\mathcal{Y}$.

Similarly, we set $\frac{\partial \hat{l}(\theta)}{\partial \sigma_x^2}$ to $0$. We first define the empirical variance of $\mathcal{X}$ as $\hat{\sigma}_x^2 = \frac{1}{N} \sum_{x \in \mathcal{X}} (x-\mu_x)^2$. By using the complementary slackness condition and the fact that $\frac{\partial \hat{l}(\theta)}{\partial \sigma_u^2}=0$, we achieve
\begin{align}
    &\frac{\partial \hat{l}(\theta)}{\partial \sigma_x^2} \doteq 0 \\
    \iff& 
    \frac{1}{2\sigma_x^2} 
    -\frac{1}{2} \frac{\hat{\sigma}_x^2}{\sigma_x^{4}}
    -\lambda_x + \lambda_u
    = 0 \\
    \iff& 
    \sigma_x^2
    -\hat{\sigma}_x^2
    +2\lambda_u\sigma_x^4
    = 0 \\
    \iff& 
    \sigma_x^2
    = \hat{\sigma}_x^2
    -2\lambda_u\sigma_x^4 \,.
    \label{eq:var_x_deriv}
\end{align}
Finally, the complementary slackness condition leaves us with two cases to consider: (1) $\lambda_u=0$ and (2) $\sigma_u^2=0$. In the former case, it directly follows from (\ref{eq:var_u_deriv}) and then (\ref{eq:var_x_deriv}) that 
\begin{align}
	\text{Case 1:} \qquad & \text{valid iff.} \; \hat{\sigma}_y^2 \geq \hat{\sigma}_x^2 \\
    &\sigma_x^2 = \hat{\sigma}_x^2 \\
    &\sigma_u^2 = \hat{\sigma}_y^2 - \hat{\sigma}_x^2 \,.
\end{align}
In the case of $\sigma_u^2=0$, we first obtain from (\ref{eq:var_u_deriv}) that
\begin{gather}
    2 \lambda_u \sigma_x^4  = \sigma_x^2 - \hat{\sigma}_y^2 \,.
\end{gather}
Inserting this into (\ref{eq:var_x_deriv}) gives the desired solution for $\sigma_x^2$ as
\begin{align}
	\text{Case 2:} \qquad & \text{valid iff.} \; \hat{\sigma}_y^2 \leq \hat{\sigma}_x^2 \\
	&\sigma_x^2 = \frac{\hat{\sigma}_x^2 + \hat{\sigma}_y^2}{2} \\
	&\sigma_u^2 = 0 \,.
\end{align}
The second case thus corresponds to the solution where $u$ is an unknown constant variable.

\section{Closed-Form Solution for the 1-Dimensional Gaussian Case using DeFlow with a Single Affine Layer}\label{supp:restricted-base}
In our proposed DeFlow method, we restrict the base distribution $p_x$ to be $\mathcal{N}(0,1)$, while keeping $p_u=\mathcal{N}(\mu_u,\sigma_u^2)$. We show that a single-affine-layer flow $f_{\theta}(x)=ax+b$ is able to obtain the an optimal solution for the 1-dimensional Gaussian setting from the previous section under this restriction. 
To do so, we simply set 
\begin{align}
    a=\frac{1}{\sigma_x} \,, \quad b=-\frac{\mu_x}{\sigma_x}
\end{align}
where $\mu_x$ and $\sigma_x$ are the optimal estimates obtained in the previous section. Intuitively, we can interpret the single-layer flow as a learned normalization layer, that ensures a standard normal distribution in the latent space.
To recover the optimal parameters $\Tilde{\mu}_u^2$ and $\Tilde{\sigma}_u^2$ of $p_u$, we need to adjust the optimal values retrieved in the previous section accordingly to this normalization and obtain
\begin{align}
    \Tilde{\mu}_u = \frac{\mu_u}{\sigma_x}
    \,, \quad
    \Tilde{\sigma}_u^2 = \frac{\sigma_u^2}{\sigma_x^2} \,.
\end{align}

This shows that the restriction of $p_x$ to be standard normal simply leads to an absorption of the required normalization in an affine layer of the flow model.

\section{Derivation of the Domain Invariant Conditional DeFlow Method} \label{supp:cond-flow}
To generalize the formulation of DeFlow from Sec.~\ref{sec:flow}
to include the domain invariant conditioning $h(x)$, we extend the flow network to $z_{x|h(x)} = f_\theta(x; h(x))$ and $z_{y|h(y)} = f_\theta(y; h(y))$. 
By invertibility in the first arguments of $f_\theta$, samples can then be retrieved by
\begin{subequations}
\label{eq:model-uncond-supp}
\begin{align}
    &x = f^{-1}_\theta(z_x;h(x)) \,,\quad y = f^{-1}_\theta(z_x + u;h(y)) \\
    &z_{x|h(x)} \sim \norm(0,I) \,,\quad u \sim p_u = \norm(\mu_u, \Sigma_u) \,,\quad z_{x|h(x)} \perp u \,.
\end{align}
\end{subequations}
Then, by domain invariance $h(x)=h(y)$, it follows that we can sample from the conditional distribution $p(y|x,h(x),h(y))=p(y|x)$ using
\begin{equation}
\label{eq:cond-sample-supp}
    y = f^{-1}_\theta(f_\theta(x;h(x)) + u;h(x)) \sim p(y|x)\\ \,
\end{equation}
where $u \sim \norm(\mu_u, \Sigma_u)$.

By the change of variables formula, we obtain the differentiable expressions for the conditional marginal distributions,
\begin{subequations}
\label{eq:marginal-density-cond}
\begin{align}
    p(x|h(x)) &= \big|\det D f_\theta(x;h(x))\big| \cdot \norm(f_{\theta}(x;h(x)); 0, I) \\
    p(y|h(y)) &= \big|\det D f_\theta(y;h(y))\big| \cdot  \norm(f_{\theta}(y;h(y)); \mu_u, I+\Sigma_u) \,.
\end{align}
\end{subequations}
As in the unconditional case, the first factor is given by the determinant of the Jacobian $D f_\theta$ of the flow network, while the second factor stems from the Gaussian base distributions from out latent space formulation.

We can then use \eqref{eq:marginal-density-cond} to allow the optimization of the new negative log-conditional-likelihood objective
\begin{equation}
    L(\theta) =  - \frac{1}{n} \sum_{i=1}^n \ln p_x(x_i| h(x_i)) - \frac{1}{m} \sum_{j=1}^m \ln p_y(y_j|h(y_j) \,.
\end{equation}

\section{DeFlow Degradation Results} \label{supp:DeFlow-results}

\begin{figure*}
    \centering
    \resizebox{1.00\linewidth}{!}{%
    \includegraphics{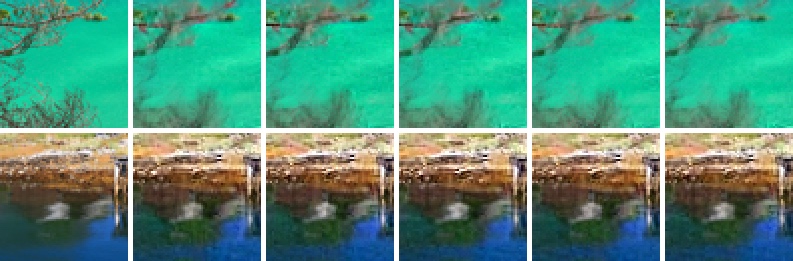}
    }\\%

    \resizebox{1.00\linewidth}{!}{%
    \includegraphics{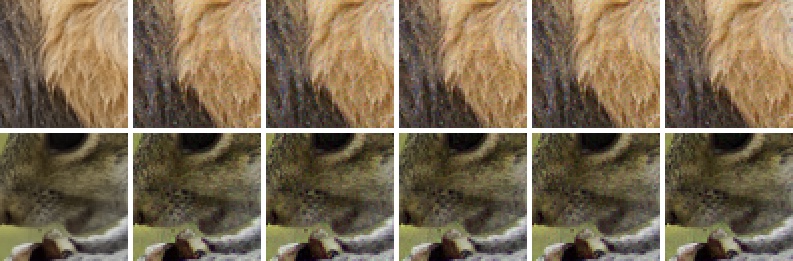}
    }\\%
    \resizebox{1.00\linewidth}{!}{%
    \begin{tabular}{C{3cm} C{15cm}}
            Clean Input &  Different Samples with $\tau=1.0$
    \end{tabular}
    }
    
    \caption{Multiple degraded samples of a clean input image (left column) using DeFlow on the AIM-RWSR (top two rows) and NTIRE-RWSR (bottom two rows). }
    \label{fig:stochastic_supp}
\end{figure*}

\parsection{Stochasticity of Degradtations}
Current GAN based approaches \cite{Fritsche19Frequency, Ji_2020_CVPR_Workshops, wei2020unsupervised, AIM2019RWSRchallenge} model the degradation process as a deterministic mapping, ignoring its inherent stochastic nature. In contrast, DeFlow learns the conditional distribution $p(y|x)$ of a degraded image $y$ given a clean image $x$ and thereby allows sampling multiple degraded versions of a single clean image. As shown in Fig.~\ref{fig:stochastic_supp}, different degraded samples from DeFlow feature different yet realistic noise characteristics without noticeable bias or recurring patterns. 

\begin{figure*}
    \centering
    \resizebox{1.00\linewidth}{!}{%
    \includegraphics{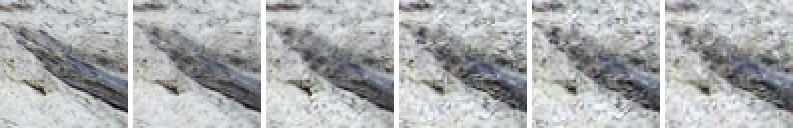}
    }\\%

    \resizebox{1.00\linewidth}{!}{%
    \includegraphics{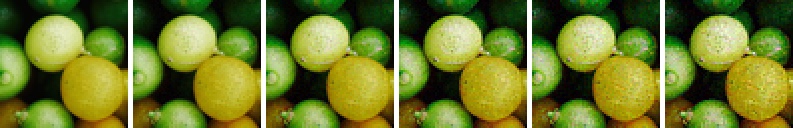}
    }\\%
    \resizebox{1.00\linewidth}{!}{%
    \begin{tabular}{C{3cm} C{3cm} C{3cm} C{3cm} C{3cm} C{3cm}}
            Clean Input & $\tau=0.33$  &  $\tau=0.66$ & $\tau=1.0$ & $\tau=1.33$ & $\tau=1.66$
    \end{tabular}
    }
    
    \caption{Sampling degradations from DeFlow with increasing temperature $\tau$ in \eqref{eq:heat-sample} on the AIM-RWSR (top row) and NTIRE-RWSR (bottom row). }
    \label{fig:heat_supp}
\end{figure*}

\parsection{Varying Degradation Strength}
We further show that DeFlow can be extended to enable sampling degradations at different strengths. To do so, we include a temperature parameter $\tau$ that scales the sampled shift-vector $u$ in the latent space. This extends (8) 
to 
\begin{equation}
\label{eq:heat-sample}
    y = f^{-1}_\theta\big(f_\theta(x;h(x)) + \tau u;\,h(x)\big).
\end{equation}
As shown in Figure \ref{fig:heat_supp}, setting $\tau < 1$ yields more nuanced degradations, while $\tau > 1$ amplifies the noise. 

\section{Visual Comparison} \label{supp:visual-comp}

While we compared DeFlow to current methods using reference and no-reference based evaluation metrics and a user study, we here provide detailed visual results.

\begin{figure*}[t]
    \centering
    \resizebox{1.00\linewidth}{!}{%
    \includegraphics{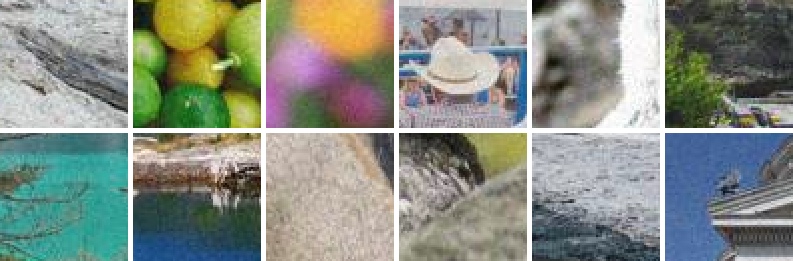} 
    }\\%
    
    \caption{AIM-RWSR: examples of noisy image patches.}
    \label{fig:AIM_noisy_supp}
\end{figure*}

\begin{figure*}
    \centering
    \resizebox{1.00\linewidth}{!}{%
    \includegraphics{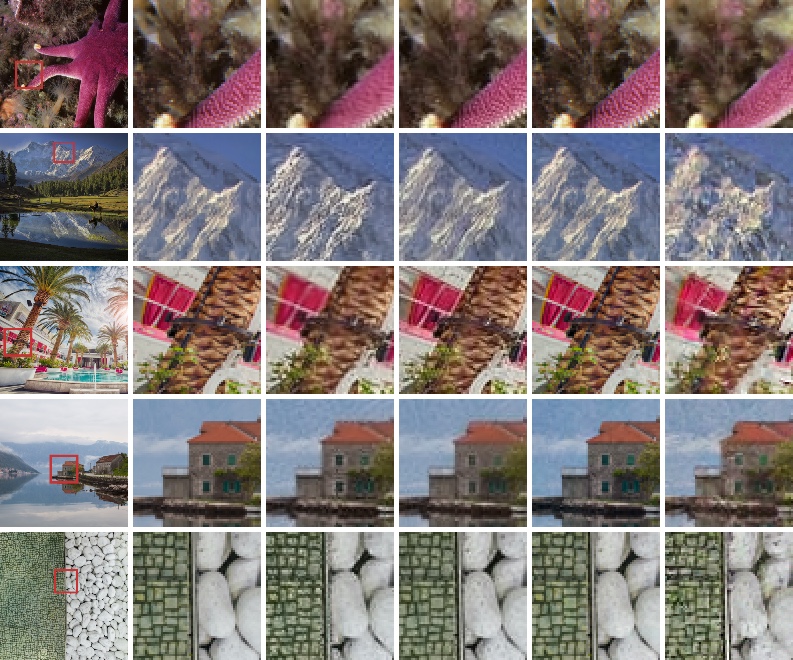} 
    }\\%
    \vspace{-4pt}
    \resizebox{1.00\linewidth}{!}{%
    \begin{tabular}{C{4cm} C{4cm} C{4cm} C{4cm} C{4cm} C{4cm}}
            & Clean Input  &  DASR \cite{wei2020unsupervised}& Frequency Separation \cite{Fritsche19Frequency} & Impressionism \cite{Ji_2020_CVPR_Workshops} & \textbf{DeFlow} (ours) 
    \end{tabular}
    }%
    \caption{AIM-RWSR: examples of clean inputs and corresponding synthetically degraded versions from different domain adaption methods.}
    \label{fig:AIM_DEG_supp}
\end{figure*}

\begin{figure*}
    \centering
    \resizebox{1.00\linewidth}{!}{%
    \includegraphics{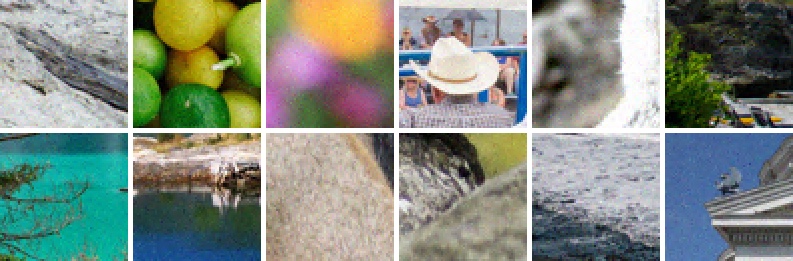} 
    }%
    \caption{NTIRE-RWSR: examples of noisy image patches.}
    \label{fig:NTIRE_noisy_supp}
\end{figure*}

\begin{figure*}
    \centering
    \resizebox{1.00\linewidth}{!}{%
    \includegraphics{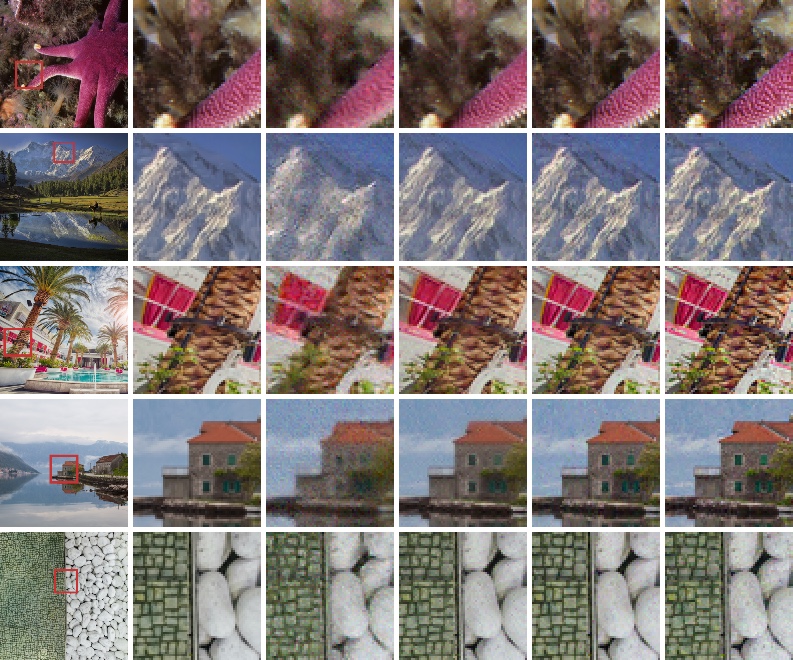} 
    }\\%
    \vspace{-4pt}
    \resizebox{1.00\linewidth}{!}{%
    \begin{tabular}{C{4cm} C{4cm} C{4cm} C{4cm} C{4cm} C{4cm}}
            & Clean Input  &  CycleGAN \cite{AIM2019RWSRchallenge}  & Frequency Separation \cite{Fritsche19Frequency} & Impressionism \cite{Ji_2020_CVPR_Workshops} & \textbf{DeFlow} (ours) 
    \end{tabular}
    }%
    \caption{NTIRE-RWSR: examples of clean inputs and corresponding synthetically degraded versions from different domain adaption methods. }
    \label{fig:NTIRE_DEG_supp}
\end{figure*}

\begin{figure*}
    \centering
    \resizebox{1.00\linewidth}{!}{%
    \includegraphics{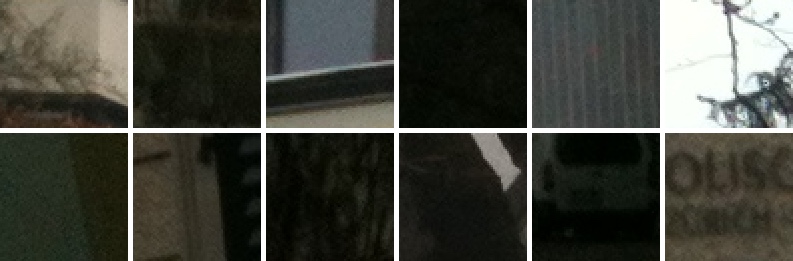} 
    }%
    \caption{DPED-RWSR: examples of noisy image patches.}
    \label{fig:DPED_noisy_supp}
\end{figure*}

\begin{figure*}
    \centering
    \resizebox{0.833\linewidth}{!}{%
    \includegraphics{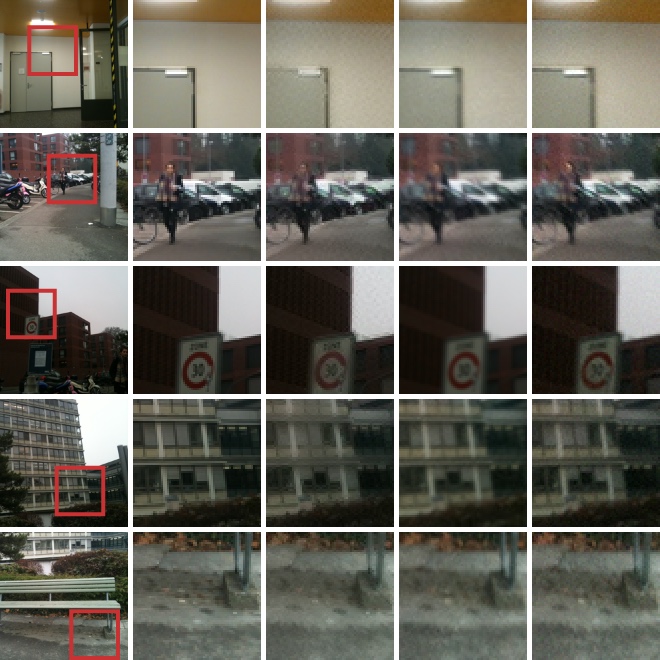} 
    }\\%
    \vspace{-4pt}
    \resizebox{0.833\linewidth}{!}{%
    \begin{tabular}{C{4cm} C{4cm} C{4cm} C{4cm} C{4cm} C{4cm}}
            & Clean Input   \cite{AIM2019RWSRchallenge} & Frequency Separation \cite{Fritsche19Frequency} & Impressionism \cite{Ji_2020_CVPR_Workshops} & \textbf{DeFlow} (ours) 
    \end{tabular}
    }%
    \caption{DPED RWSR: examples of clean inputs and corresponding synthetically degraded versions from different domain adaption methods.  Note, that we did not include CycleGAN \cite{AIM2019RWSRchallenge} as differing to the other approaches it is trained to degrade images from DIV2k with DPED noise instead of down-sampled DPED images. }
    \label{fig:DPED_DEG_supp}
\end{figure*}

\noindent\parsection{Degradation Results:} 
We thus show examples of the synthetic degradations generated from different methods in Figures \ref{fig:AIM_DEG_supp}, \ref{fig:NTIRE_DEG_supp}, and \ref{fig:DPED_DEG_supp} for the AIM-, NTIRE-, and DPED-RWSR datasets. As a reference, we further provide examples of real noisy image patches from the respective datasets in Figures \ref{fig:AIM_noisy_supp}, \ref{fig:NTIRE_noisy_supp}, and \ref{fig:DPED_noisy_supp}. We notice that DeFlow consistently adds more noise compared to the other methods. Yet, on all datasets, the degradations from DeFlow resemble the real noisy data, whereas other learned methods struggle to pickup on the noise characteristics.

\begin{figure*}
\centering
\resizebox{1.00\linewidth}{!}{%
\includegraphics{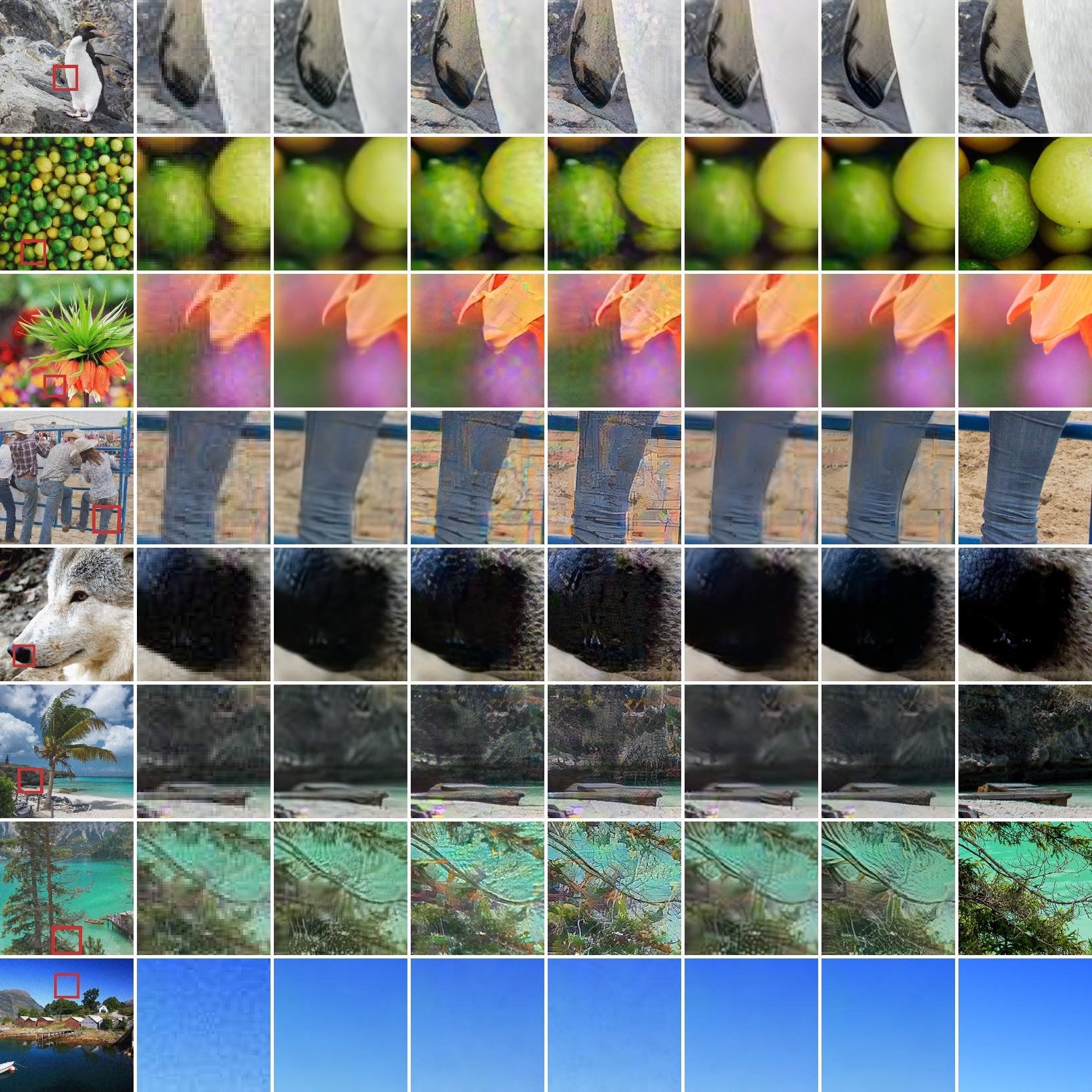} 
}\\%
\vspace{-4pt}
\resizebox{1.005\linewidth}{!}{%
\begin{tabular}{C{3.5cm} C{3.5cm} C{3.5cm} C{3.5cm} C{3.8cm} C{3.5cm} C{3.5cm} C{3.5cm}}
        {} & LR  & White Noise $\sigma\!=\!0.04$ & DASR$^\dagger$ \cite{wei2020unsupervised}& Frequency Separation$^\dagger$ \cite{Fritsche19Frequency} & Impressionism$^\dagger$ \cite{Ji_2020_CVPR_Workshops} & \textbf{DeFlow} (ours) & GT
\end{tabular}
}%
\caption{AIM-RWSR: Super-resolution results on the validation set. Methods marked with $\dagger$ employ the same SR pipeline as DeFlow and the baselines. Crops were chosen at random for an unbiased comparison.}
\label{fig:AIM_SR_supp}
\end{figure*}

\begin{figure*}
\centering
\resizebox{1.00\linewidth}{!}{%
\includegraphics{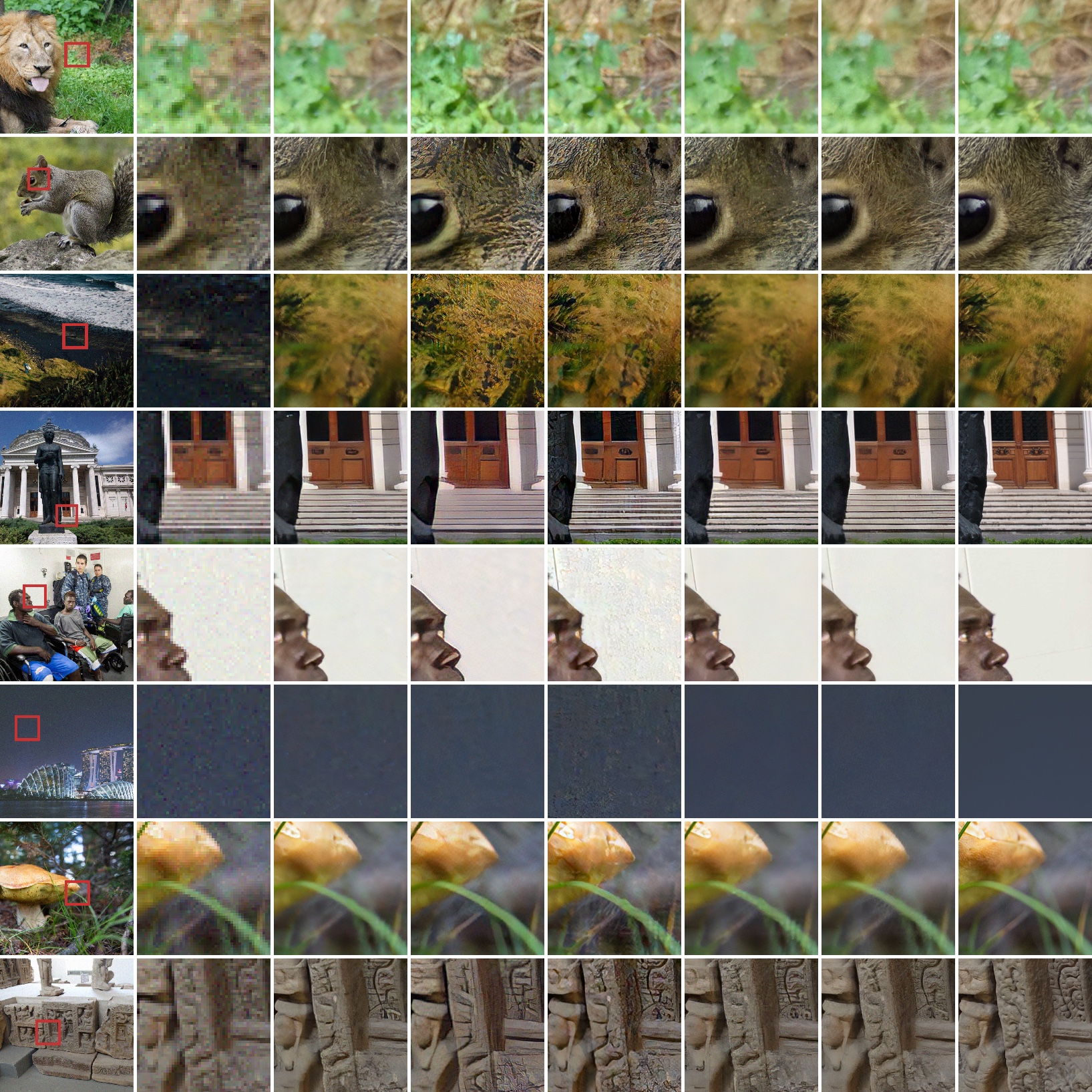} 
}\\%
\vspace{-4pt}
\resizebox{1.005\linewidth}{!}{%
\begin{tabular}{C{3.5cm} C{3.5cm} C{3.9cm} C{3.5cm} C{3.9cm} C{3.5cm} C{3.5cm} C{3.5cm}}
        {} & LR  & White Noise $\sigma\!\sim\!\mathcal{U}(0,0.06)$~ & DASR$^\dagger$ \cite{wei2020unsupervised}~~~& Frequency Separation$^\dagger$\cite{Fritsche19Frequency}~~~~ & Impressionism$^\dagger$ \cite{Ji_2020_CVPR_Workshops}~~~~~~ & \textbf{DeFlow} (ours)~~~~ & GT
\end{tabular}
}%
\caption{NTIRE-RWSR: Super-resolution results on the validation set. Methods marked with $\dagger$ employ the same SR pipeline as DeFlow and the baselines. Crops were chosen at random for an unbiased comparison.}
\label{fig:NTIRE_SR_supp}
\end{figure*}

\begin{figure*}
\centering
\resizebox{1.00\linewidth}{!}{%
\includegraphics{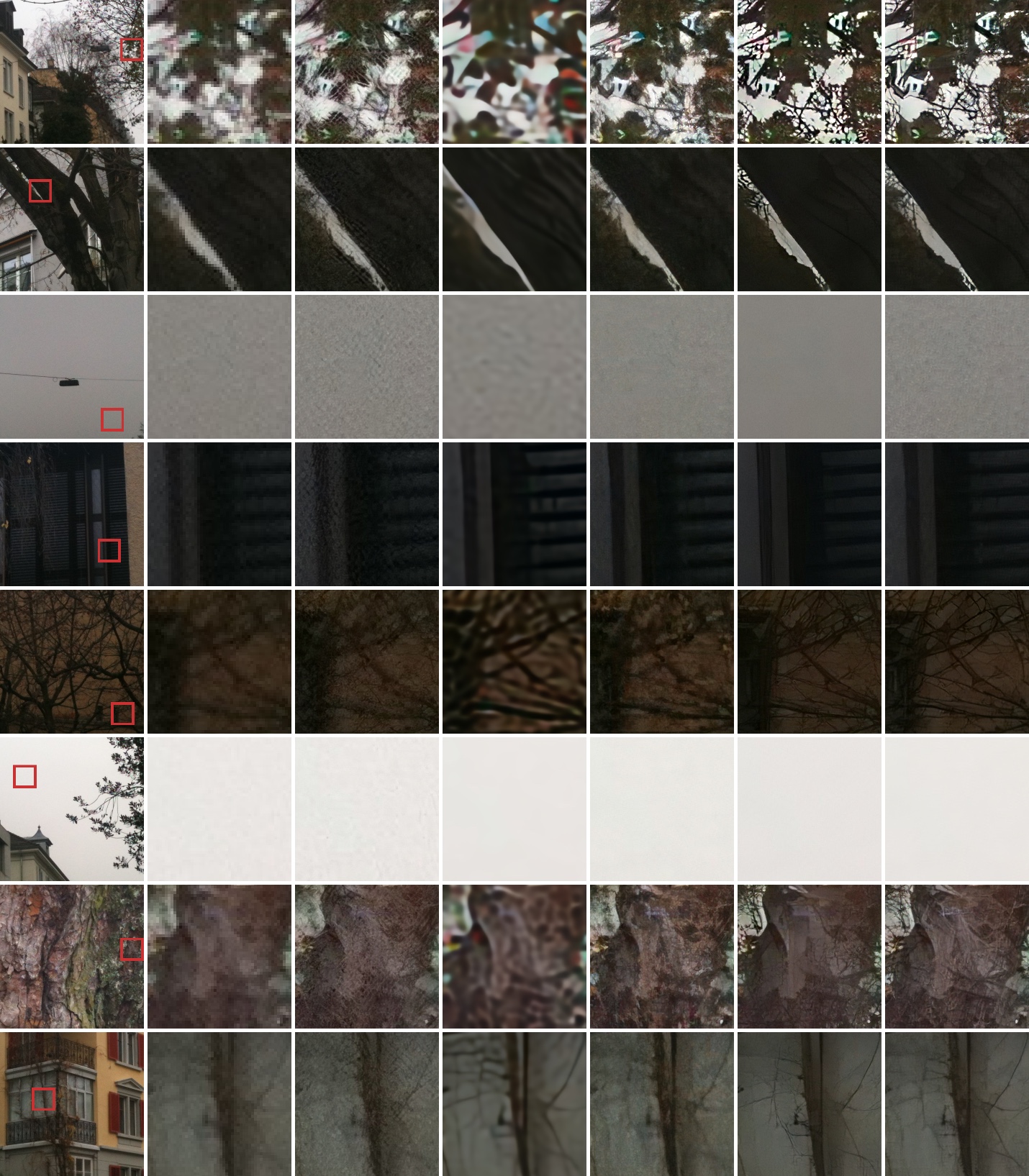} 
}\\%
\vspace{-4pt}
\resizebox{1.005\linewidth}{!}{%
\begin{tabular}{C{3.5cm} C{3.5cm} C{3.5cm} C{3.5cm} C{3.8cm} C{3.5cm} C{3.5cm}}
        {} & LR  & No Degradation \cite{wang2018esrgan} & CycleGAN \cite{AIM2019RWSRchallenge} & Frequency Separation$^\dagger$ \cite{Fritsche19Frequency} & Impressionism \cite{Ji_2020_CVPR_Workshops} & \textbf{DeFlow} (ours) 
\end{tabular}
}
\caption{DPED-RWSR: Super-resolution results on the validation set. Ground truth reference images do not exist for this dataset as it consists of real-world low-quality smartphone photos. Methods marked with $\dagger$ employ the same SR pipeline as DeFlow and the baselines. Crops were chosen at random for an unbiased comparison.}
\label{fig:DPED_SR_supp}
\end{figure*}

\noindent\parsection{Real-World Super-Resolution Performance:}
Further, we provide results of the downstream real-world super-resolution task of the different methods on the AIM-, NTIRE-, and DPED-RWSR datasets in Figures 
\ref{fig:AIM_SR_supp},
\ref{fig:NTIRE_SR_supp}, and 
\ref{fig:DPED_SR_supp},
respectively. 
It is noticeable, that our proposed approach introduces fewer artifacts than the other methods across all datasets. Further, DeFlow is able to reconstruct fine details and provides sharper images than the White Noise model, which performs surprisingly well on the synthetic datasets. On DPED, the performance of the DeFlow degradations is comparable to the handcrafted approach of Impressionism \cite{Ji_2020_CVPR_Workshops}. While DeFlow retains more noise in smooth patches, Impressionism tends to over-smooth textures.

\section{Details of the User Study} \label{supp:user-study}
In this section, we give insight into how we conducted the user study. 
On AIM and DPED we chose the top 7 models by their LPIPS score to compare in the user study. 
On DPED we decided to only compare against Frequency Separation \cite{Fritsche19Frequency} and Impressionism \cite{Ji_2020_CVPR_Workshops} both with their super-resolution pipeline and ours, as we found that other methods performed considerably worse.

For all datasets we used the following set-up for the user study: Participants were shown the same random crop from two different super-resolution models. In addition, we showed them the whole image where the cropped patch was marked in red. Participants were then asked to pick the super-resolved patch that looks more realistic. For that we used three random crops of size $80 \times 80$ pixels per image of each validation dataset and asked five different study participants per pair.


\typeout{get arXiv to do 4 passes: Label(s) may have changed. Rerun}
\end{document}